%% file: 2025_naacl_qgqa.tex
\pdfoutput=1
\documentclass[11pt]{article}
\usepackage{float}

\input{style/preamble}

\usepackage[utf8]{inputenc}
\usepackage{pgfplots}
\usepackage{inconsolata}
\usepackage{booktabs}
\usepackage{multirow}
\usepackage{amsmath}
\usepackage{amsfonts}
\usepackage{amsthm}
\usepackage{pifont}
\usepackage{xspace}
\usepackage{graphicx}
\usepackage{bm}
\usepackage{xcolor, colortbl}
\usepackage[most]{tcolorbox}
\usepackage{csquotes}
\usepackage{anyfontsize}
\usepackage{comment}
\usepackage{float}
\usepackage{cuted}
\usepackage{tikz}
\usepackage{listings,multicol}
\usepackage{filecontents}
\usepackage{dsfont}
\DeclareUnicodeCharacter{2212}{−}
\usepgfplotslibrary{groupplots,dateplot}
\usetikzlibrary{patterns,shapes.arrows}
\pgfplotsset{compat=newest}

\usepackage{tikzscale}
\usepackage{amsmath}
\newcommand{\probP}{\text{I\kern-0.15em P}}

\usepackage{multirow, colortbl}

\usepackage{tabularx,booktabs}
\usepackage{makecell}

\usepackage[normalem]{ulem}
\useunder{\uline}{\ul}{}

\usepackage{graphicx}

\definecolor{ablation6}{HTML}{fcefed}
\definecolor{ablation_tie}{HTML}{fce3e1}

\definecolor{ablation5}{HTML}{fcd8d4}
\definecolor{ablation4}{HTML}{FBC3BC}
\definecolor{ablation3}{HTML}{F7A399}
\definecolor{ablation2}{HTML}{F38375}
\definecolor{ablation1}{HTML}{EF6351}
\definecolor{UMDred}{HTML}{ed1c24}
\definecolor{UMDyellow}{HTML}{ffc20e}
\definecolor{CustomGreen}{HTML}{1FC801}

\usepackage[]{acl2023}

\usepackage{xspace}
\usepackage{cancel}
\usepackage[utf8]{inputenc}
\usepackage{pgfplots}
\usepackage{dsfont}
\DeclareUnicodeCharacter{2212}{−}
\usepgfplotslibrary{groupplots,dateplot}
\usetikzlibrary{patterns,shapes.arrows}
\pgfplotsset{compat=newest}

\newcommand{\fwd}{\textsc{QA}\xspace}
\newcommand{\bwd}{\textsc{RQA}\xspace}

\newcommand{\settingName}{domains\xspace}

\definecolor{bggray}{rgb}{0.95, 0.95, 0.95}
\usepackage[%
    framemethod=tikz,
    skipbelow=\topskip,
    skipabove=\topskip
]{mdframed}
\mdfsetup{%
    leftmargin=0pt,
    rightmargin=0pt,
    backgroundcolor=bggray,
    middlelinecolor=black,
    roundcorner=3
}

\newtcolorbox[list inside=prompt,auto counter,number within=section]{prompt}[1][]{
    colbacktitle=black!60,
    fonttitle=\small,
    coltitle=white,
    fontupper=\footnotesize,
    boxsep=4pt,
    left=0pt,
    right=0pt,
    top=0pt,
    bottom=0pt,
    boxrule=1pt,
    width=\textwidth, % Ensure the box takes the full column width
    enlarge left by=0mm, % Ensure no additional margins on the left
    enlarge right by=0mm, % Ensure no additional margins on the right
    #1,
}

\usepackage[%
    framemethod=tikz,
    skipbelow=\topskip,
    skipabove=\topskip
]{mdframed}
\mdfsetup{%
    leftmargin=0pt,
    rightmargin=0pt,
    backgroundcolor=white,
    middlelinecolor=black,
    roundcorner=3
}

\newtcolorbox[list inside=prompt,auto counter,number within=section]{summary}[1][]{
    colbacktitle=blue!10,
    fonttitle=\small,
    coltitle=black,
    fontupper=\footnotesize,
    boxsep=4pt,
    left=0pt,
    right=0pt,
    top=0pt,
    bottom=0pt,
    boxrule=1pt,
    width=\textwidth, % Ensure the box takes the full column width
    enlarge left by=0mm, % Ensure no additional margins on the left
    enlarge right by=0mm, % Ensure no additional margins on the right
    #1,
}

\DeclareUnicodeCharacter{2318}{\applecmd}

\definecolor{ablation6}{HTML}{fcefed}
\definecolor{ablation_tie}{HTML}{fce3e1}

\definecolor{ablation5}{HTML}{fcd8d4}
\definecolor{ablation4}{HTML}{FBC3BC}
\definecolor{ablation3}{HTML}{F7A399}
\definecolor{ablation2}{HTML}{F38375}
\definecolor{ablation1}{HTML}{EF6351}
\definecolor{yellowcolor}{HTML}{ffc20e}
\definecolor{purplecolor}{HTML}{e598fa}
\definecolor{bluecolor}{HTML}{8cd2f5}
\definecolor{UMDred}{HTML}{ed1c24}
\definecolor{UMDyellow}{HTML}{ffc20e}
\definecolor{CustomGreen}{HTML}{1FC801}

\newcommand{\MyColorBoxYellow}[2][yellowcolor]%
{%
    \settowidth{\Width}{#2}%
    \colorbox{#1}%
    {%      
        \raisebox{-\DepthReference}%
        {%
                \parbox[b][\HeightReference+\DepthReference][c]{\Width}{\centering#2}%
        }%
    }%
}

\newcommand{\MyColorBoxPurple}[2][purplecolor]%
{%
    \settowidth{\Width}{#2}%
    \colorbox{#1}%
    {%      
        \raisebox{-\DepthReference}%
        {%
                \parbox[b][\HeightReference+\DepthReference][c]{\Width}{\centering#2}%
        }%
    }%
}

\usepackage[]{algpseudocode}
\usepackage[]{algorithm}
\usepackage{float}
\algtext*{EndFor}%
\algtext*{EndProcedure}%

\usepackage{booktabs}
\usepackage[normalem]{ulem}
\useunder{\uline}{\ul}{}

\usepackage{times}
\usepackage{latexsym}
\usepackage{adjustbox}

\usepackage[T1]{fontenc}
\usepackage{amsmath}
\usepackage{amssymb}
\usepackage{booktabs}
\usepackage{multirow}
\usepackage{scalerel,xparse}
\usepackage[utf8]{inputenc}
\usepackage{cleveref}
\usepackage{microtype}
\usepackage{enumitem}
\crefformat{section}{\S#2#1#3}
\crefformat{subsection}{\S#2#1#3}
\crefformat{subsubsection}{\S#2#1#3}

\title{Reverse Question Answering:\\Can an LLM Write a Question so Hard (or Bad) that it Can't Answer?}

\author{Nishant Balepur$^{1}$ \hspace{0.5cm} Feng Gu$^{1}$  \hspace{0.5cm} \textbf{Abhilasha Ravichander}$^{2}$ \hspace{0.5cm}  \textbf{Shi Feng}$^{3}$ \\ \textbf{Jordan Boyd-Graber}$^{1}$ \hspace{0.5cm} \textbf{Rachel Rudinger}$^{1}$ \\
  $^{1}$University of Maryland \hspace{0.5cm}
    $^{2}$University of Washington \\
  $^{3}$George Washington University \\ 
  \texttt{\{nbalepur, rudinger\}@umd.edu} \hspace{0.5cm} \texttt{jbg@.umiacs.umd.edu}
}

\begin{document}
\maketitle

\input{2025_naacl_qgqa/sections/00_abstract}

\input{2025_naacl_qgqa/sections/10_intro}

\input{2025_naacl_qgqa/sections/20_problem}

\input{2025_naacl_qgqa/sections/30_experimental_setup}

\input{2025_naacl_qgqa/sections/40_results}

\input{2025_naacl_qgqa/sections/50_related_work}

\input{2025_naacl_qgqa/sections/60_conclusion}

\input{2025_naacl_qgqa/sections/70_limitations_ethics}

% \jbgcomment{This "balepur" guy is cited a little too often.  Tone it down for submission, can be added back in for camera ready.}

\bibliography{custom}
\bibliographystyle{acl_natbib}

\appendix \label{sec:appendix}

\clearpage

\input{2025_naacl_qgqa/sections/80_appendix}

\end{document}

%% file: style/preamble.tex
\usepackage[a-1b]{pdfx}

\usepackage{framed}
\usepackage{mdwlist}
\usepackage{siunitx}
\usepackage{latexsym}
\usepackage{colortbl}
\usepackage{xcolor}
\usepackage{nicefrac}
\usepackage{booktabs}
\usepackage{fnpct}
\usepackage{amsfonts}
\usepackage[T1]{fontenc}
\usepackage{bold-extra}
\usepackage{amsmath}
\usepackage{amssymb}
\usepackage{bm}
\usepackage{graphicx}
\usepackage{mathtools}
\usepackage{microtype}
\usepackage{multirow}
\usepackage{multicol}
\usepackage{xpatch}
\usepackage{latexsym,comment}
\usepackage[normalem]{ulem}

\newcommand*{\missingreference}{{\Huge \colorbox{red}{?reference?}}}
\newcommand*{\missingcitation}{{\Huge \colorbox{red}{?citation?}}}

\makeatletter
\xpatchcmd{\@setref}{\bfseries}{\missingreference}{}{}
\def\@citex[#1]#2{\leavevmode
    \let\@citea\@empty
    \@cite{\@for\@citeb:=#2\do
        {\@citea\def\@citea{,\penalty\@m\ }%
            \edef\@citeb{\expandafter\@firstofone\@citeb\@empty}%
            \if@filesw\immediate\write\@auxout{\string\citation{\@citeb}}\fi
            \@ifundefined{b@\@citeb}{\hbox{\reset@font\missingcitation}%
                \G@refundefinedtrue
                \@latex@warning
                {Citation `\@citeb' on page \thepage \space undefined}}%
            {\@cite@ofmt{\csname b@\@citeb\endcsname}}}}{#1}}
\makeatother

\newcommand{\mm}[0]{\abr{llm}}

\newcommand{\gem}[1]{\mbox{\textsc{gem}}}
\newcommand{\abr}[1]{\textsc{#1}}

\newcommand{\hidetext}[1]{}
\newcommand{\ignore}[1]{}
\newif\ifcomment
\commenttrue

\ifcomment
    \newcommand{\pinaforecomment}[3]{\colorbox{#1}{\parbox{.8\linewidth}{#2: #3}}}

    \newcommand{\prtodo}[1]{\pinaforecomment{lightblue}{pr}{#1}}
    \newcommand{\prtodoi}[1]{\pinaforecomment{lightblue}{pr}{#1}}
\else
    \newcommand{\pinaforecomment}[3]{}
    \newcommand{\prtodo}[1]{}
    \newcommand{\prtodoi}[1]{}
\fi

\newcommand{\smallurl}[1]{ \begin{tiny}\url{#1}\end{tiny}}

\definecolor{lightblue}{HTML}{3cc7ea}
\definecolor{CUgold}{HTML}{CFB87C}
\definecolor{grey}{rgb}{0.95,0.95,0.95}
\definecolor{ceil}{rgb}{0.57, 0.63, 0.81}
\definecolor{UMDred}{HTML}{ed1c24}
\definecolor{UMDyellow}{HTML}{ffc20e}

\newcommand{\qb}[0]{\abr{qb}}

\newcommand{\nlp}[0]{\abr{nlp}}

%% file: 2025_naacl_qgqa/sections/00_abstract.tex
% \jbgcomment{Brainstorming titles:

% Answer Questioning: Can \abr{ai} generate a question so good/bad \abr{ai}s cannot answer it?

% }

\begin{abstract} {
% \ar{Can we very briefly define abductive and deductive in the abstract}

Question answering (\fwd)---giving correct answers to
questions---is a popular task, but we test \textbf{\textit{reverse}
  question answering (\bwd)}: for an input answer, give a
question with that answer.
Past work tests \fwd and \bwd separately,
but we test them jointly, comparing their difficulty, aiding benchmark
design, and checking reasoning consistency.
We run 16~\mm{}s on \fwd and
\bwd with trivia questions/answers, revealing:
1) Versus \fwd, \mm{}s
are much less accurate in \bwd for numerical answers, but~slightly
more accurate in \bwd for textual answers;
2) \mm{}s often answer
their own invalid questions from \bwd accurately in \fwd, so \bwd
errors are not from knowledge gaps alone;
3) \bwd errors correlate
with question difficulty and inversely correlate with answer
frequencies in the Dolma corpus; and
4) \mm{}s struggle to provide valid
multi-hop questions.
By finding question and answer types that lead to
\bwd errors, we suggest improvements for \mm{}
reasoning.
\footnote{Code and data available at \url{https://github.com/nbalepur/Reverse-QA}}
% \footnote{\url{http:/anonymous.4open.science/r/aq-qa}}

% Large language models (\mm{}s) use \textit{deduction}, concluding from premises, and \textit{abduction}, inferring explanations, in zero-shot settings, but the consistency of these reasoning strategies is underexplored.
% We probe 16 \mm{}s to accurately answer questions (\abr{qa}, deduction) and give questions about entities (QG, abduction) using trivia question/answer pairs.
% We find: 1) Compared to deduction, \mm{}s are much less accurate in numerical abduction but more accurate in text-based abduction;
% 2) Consistency checks, where \mm{}s answer their own questions, show that \mm{} numerical abduction is so weak that models can often detect their own question inaccuracies; and
% 3) QG failures correlate with pretraining token count and question difficulty, and \mm{}s struggle to generate multi-hop questions.
% 
}
% \ar{can we make this more concrete: by identifying systematic failures etc}
\end{abstract}

%% file: 2025_naacl_qgqa/sections/10_intro.tex
\section{Reversing the Question Answering Task}

Question answering (\fwd) is a long-standing task in
\nlp~\cite{green1961baseball}.
For an input question~$q$, \fwd deduces the correct answer
$a$~\cite{reiter1989deductive}.
More recently, large language models (\mm{}s) do the reverse---given
an answer~$a$, generate a valid question~$q$ to which $a$ is the
answer---which we call \textbf{\textit{reverse} question answering}
(\bwd).\footnote{This definition differs from question
  generation~\cite{zhang2021review}, which grounds the answer to an input
  context.}
\bwd thus can be a part of downstream tasks like exam question
generation~\cite{biancini2024multiple} or
search query reformulation~\cite{10.1145/1718487.1718493}.

\fwd and \bwd are often tested separately, but we test them jointly,
offering two key benefits.
First, it gives insights into open questions on \mm{}
abilities, as some show \mm{}s excel in generation over comprehension
\cite[\bwd]{west2023generative}, while others claim verification is
easier \cite[\fwd]{Kadavath2022LanguageM}.
Uncovering which task is harder can guide benchmark
design~\cite{chen2024see} and inform data collection practices in
writing question-answer pairs (\cref{subsection:accuracy}; e.g., if
\bwd is~easy, get answers manually and then generate synthetic
questions).

Second, chaining \bwd and \fwd forms a consistency check for
\mm{} reasoning~\cite{liuincomplete}. \bwd---inferring just one of
many valid questions---is
\textit{abductive}~\cite{abe1998applications}, while \fwd---inferring
an answer from question premises---is
\textit{deductive}~\cite{reiter1989deductive}.
Thus, by seeing if $\fwd(\bwd(a)) \approx a$, i.e., checking if an
\mm{} can answer its own question from \bwd (Fig~\ref{fig:intro}), we
can assess \mm{}s' logical robustness in abduction and deduction
(\cref{subsection:consistency}).  This analysis can also help
determine if \mm{}s can reliably
self-verify~\cite{pan2024automatically} in downstream \bwd tasks like
writing exams~\cite{wang2018qg}.

% To reap this benefits, we conduct the first comparison of \bwd and \fwd in \mm{}s.

\input{data/data_desc}

To reap these benefits, we test if 16 \mm{}s can produce 1) questions
correctly answered by input entities~(\bwd); and 2) accurate answers
for input questions (\fwd).  We collect 3443 trivia question/answer
pairs~\cite{Rodriguez2019QuizbowlTC}, grouped by answer as either
numerical or textual entities, forming inputs to evaluate \bwd and
\fwd in varied \settingName.

\input{figures/intro}

% \jbgcomment{accuracy gap not yet defined.  Either avoid or define}
In numerical \settingName, \mm{}s are much less accurate in \bwd than
\fwd, especially integers (Fig~\ref{fig:intro}); the accuracy
difference when \mm{}s do these tasks \textbf{exceed 0.80} for
Command-R and LLaMA-3 (\cref{subsection:accuracy}).
%This suggests \mm{}s overfit on math \abr{qa} tasks, limiting
%generalization to numerical QG.
Interestingly, in textual \settingName, the trend reverses, so \mm{}s
are not consistently better generators or
validators~\cite{li2024benchmarking}.
We then design a consistency check (\cref{subsection:consistency}) to
see if \mm{}s can answer their own \bwd questions; numerical
\bwd failures are not solely due to knowledge gaps, as \mm{}s
often \textbf{answer their own invalid questions correctly} in \fwd
(33\% of cases for Claude-Opus).
We then study questions from \bwd (\cref{subsection:correlators},
\cref{subsection:error_analysis}) and find errors occur when
\mm{}s give overly-complex, multi-step questions, giving insights into
strategies---like complexity bias mitigation in preference data and
calibrating models using difficulty scores---to improve \mm{} \bwd reliability. Our contributions are:

\begin{enumerate}
    \item We use Reverse Question Answering~(\bwd) to test if \mm{}s can provide accurate questions for input answers using abductive reasoning.
    \item We reveal many \mm{}s have a surprising weakness in \bwd on numerical entities, struggling on input answers with lower pretraining token counts and when creating multi-hop questions.
    \item We design a consistency check between \bwd and \fwd, showing \mm{}s answer their own invalid questions from \bwd correctly via~\fwd.
\end{enumerate}

%% file: data/data_desc.tex
\begin{table*}[]
\small
\footnotesize
\setlength{\tabcolsep}{3.5pt}
\begin{tabular}{@{}llllc@{}}
\toprule
\textbf{Answer Type} & \textbf{Description} & \textbf{Example Question} & \textbf{Example Answer} & \textbf{Count}  \\ \midrule
(1) Number & Integers in $[100, 1000)$& What is 26 times 4? & 104 & 900  \\
(2) Number+Text & Integers with a text entity & When did Pope Hormisdas die? & 523 AD & 743 \\
(3) Easy Fact & Well-known factual entity & Who is the artist that painted Starry Night? & Vincent van Gogh & 900 \\
(4) Hard Fact & Obscure factual entity & What is the final painting by Paolo Uccello? & The Hunt in the Forest & 900 \\ \bottomrule
\end{tabular}
%\vspace{-2ex}
\caption{Description of our collected dataset for Question Answering and Reverse Questioning Answering tasks.}
\label{table:data_desc}
\end{table*}

%% file: figures/intro.tex
\begin{figure}
    \centering
    \includegraphics[width=\linewidth]{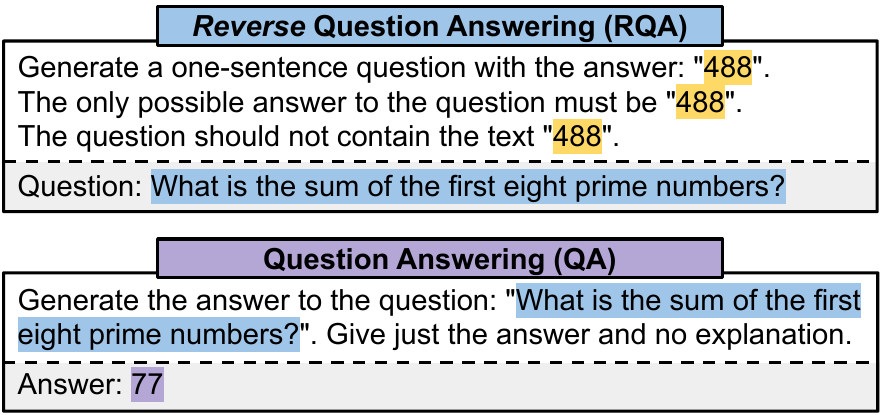}
    \vspace{-4ex}
    \setlength{\fboxsep}{0pt} % Remove the padding inside the colorbox
    \caption{\small \bwd/\fwd consistency check using GPT-4. The LLM fails to give a valid question with answer \colorbox{yellowcolor}{\strut 488} (top), but correctly gives the answer \colorbox{purplecolor}{\strut 77} for \colorbox{bluecolor}{\strut its own question} (bottom).}
    \label{fig:intro}
    %\vspace{-2ex}
\end{figure}

%% file: 2025_naacl_qgqa/sections/20_problem.tex
% \section{Problem Definition} \label{section:task}

% We test \mm{} reasoning in deduction (\textbf{\fwd}), drawing conclusions from premises \cite{li-etal-2024-cr}, and abduction (\textbf{\bwd}), inferring a likely explanation for an observation  \cite{zhao-etal-2024-uncommonsense}, via \abr{qa} and QG tasks, respectively.
% We define the tasks~below: \\
% \textbf{\noindent 1) \fwd:} Given \abr{qa} pair ($q_{true}$, $a_{true}$), the \mm{} must give an answer $a_{fwd}$ for $q_{true}$. \fwd~succeeds if $a_{true}$ matches $a_{fwd}$ semantically, since the \mm{} deductively reasoned to the correct answer of~$q$. \\
% \textbf{\noindent 2) \bwd:} Given answer $a_{true}$, the \mm{} must~give a question $q_{bwd}$ with answer $a_{true}$. \bwd succeeds if $q_{bwd}$ is correctly answered by $a_{true}$; the \mm{} abductively reasoned to one of many valid questions.

%% file: 2025_naacl_qgqa/sections/30_experimental_setup.tex
\section{Experimental Setup} \label{section:experimental_setup}

% \jbgcomment{It might worthwhile to either say how this is like \abr{qa} that everyone is used to or emphasize how it's differentitem      .}

We evaluate \mm{} abilities in question answering (\textbf{\fwd}) and
\textit{reverse} question answering (\textbf{\bwd}):  

\begin{itemize}
    \item \textbf{$\textsc{QA}(q) \rightarrow \hat{a}$}:  
    Given a question $q$ with a single answer $a$, the \mm{} produces an answer $\hat{a}$ for~$q$. \fwd{} succeeds if $a$ matches $\hat{a}$ semantically.  
    For example, given the input ``What is the name of the polygon with three sides?'' for $q$, an \mm{} using \fwd{} should give an $\hat{a}$ that matches ``triangle'' for $a$.  
    This typical \fwd{} setup tests \textit{deduction}, since the model must reason to the correct answer of $a$ based on the premises~in~$q$.

    \item \textbf{$\textsc{RQA}(a) \rightarrow \hat{q}$}:  
    Given an input answer $a$, the \mm{} must produce a question $\hat{q}$. \bwd{} succeeds if the correct answer to $\hat{q}$ is $a$ (verified via oracle, \cref{subsection:metrics}).  
    For example, given the input ``triangle'' for $a$, an \mm{} using \bwd{} could succeed with $\hat{q}$ as ``In eight-ball pool, what shape is used to rack the balls?''.  
    \bwd{} tests \textit{abduction}, as the model must reason toward one of the many valid questions with the answer~$a$.
\end{itemize}

\noindent This section describes the datasets (\cref{subsection:datasets}), models
(\cref{subsection:models}), and metrics (\cref{subsection:metrics})
used for \bwd and \fwd.

\input{data/benchmark}

% \ar{This might be a bit confusing to the reader i.e years as an example of numerical reasoning with factual knowledge}

\subsection{Dataset Collection} \label{subsection:datasets}

We study question/answer pairs~$(q, a)$ in four domains for \fwd
and \bwd inputs, based on $a$'s answer type
(Table~\ref{table:data_desc}).
We group them as numerical (Number,
Number+Text) or textual (Easy Fact, Hard Fact), providing varied
\settingName for testing.

% \jbgcomment{since we have a little more space now, perhaps refer to them as numbers plus text here so reader doesn't have to go back to table}

When $a$ is a Number, $q$ is a random, one-step math operation (what is
118+211?).
Other types are~from \abr{qanta}~\cite{Rodriguez2019QuizbowlTC}, an
expert-curated dataset of multi-sentence trivia \abr{qa} pairs.
For Easy and Hard Facts, $a$ is the answer to sampled \abr{qanta} questions, with
the last sentence\footnote{\abr{qanta} questions are paragraph-long and describe a single answer. Sentences in paragraphs are ordered in decreasing difficulty, so we use the last one, forming the easiest question. } as~$q$.
% \jbgcomment{Becaue this dataset is wierd, perhaps it would be to good to have a footnote to explain why they're multisentence and why you use the last sentence}
%
We use middle school questions for Easy Facts and college questions for Hard Facts.
We obtain Number+Text answers $a$ in \abr{qanta} by finding numbers in full questions via regex and $q$ from the sentence $a$ appears in.
One PhD student checks all \fwd pairs to ensure they are accurate (details in Appendix~\ref{appendix:dataset}).

% \jbgcomment{are you not going to define difference between easy and hard?}

% We use four types of question/answer pairs $(q, a)$ as \fwd and \bwd inputs, grouped by the answer type of $a$: Number:} All 900 integers in range $[100, 1000)$.\\
% \noindent \textbf{2) Number+Text:} 743 numbers from (1) paired with an entity (120 counties, 315 feet, 523 AD).\\
% \noindent \textbf{3) Easy:} 900 well-known factual entities~(Greece).\\
% \noindent \textbf{4) Hard:} 900 obscure factual entities~(flarf poetry).

% These types help us study \fwd and \bwd in diverse settings.
% (1) tests math reasoning, while (2) tests if \mm{}s can recall numerical facts (e.g. specific years); we refer to these settings as \textbf{numerical} \fwd/\bwd.
% Our well-known, (3), and obscure, (4), entities probe \mm{} common and long-tail factual knowledge; we refer to these as \textbf{textual}~\fwd/\bwd.

% For type (1) answers $a_{true}$, the question $q_{true}$ is a random, one-step math operation (``what is 118+211?'').
% Other types are drawn from \abr{qa}NTA \cite{Rodriguez2019QuizbowlTC}, a multi-sentence trivia \abr{qa} dataset.
% For (3) and (4), $a_{true}$ is the answer of 900 sampled middle school and college-level \abr{qa}NTA questions, respectively, with the last sentence as $q_{true}$.
% We find $a_{true}$~of type (2) in \abr{qa}NTA via regex and $q_{true}$ via the context $a_{true}$ appears in.
% We verify all 3443 ($q_{true}, a_{true}$) pairs (Appendix~\ref{appendix:dataset}).

\subsection{Models} \label{subsection:models}

We evaluate 16 \mm{}s: GPT \cite[3.5, 4, 4o]{achiam2023gpt}, Command R \cite[Command-R, Command-R+]{coherecommand}, Claude 
 \cite[Sonnet, Haiku, Opus]{anthropic2023claude}, LLaMA-3 Instruct \cite[8B, 70B]{dubey2024llama}, Yi-1.5 Chat \cite[6B, 9B, 34B]{young2024yi}, and Mistral Instruct \cite[7B, 8x7B, 8x22B]{jiang2024mixtral}. All \mm{}s use temperature~0. We list all parameters in Appendix~\ref{appendix:experimental_setup}.

The \fwd and \bwd prompts are zero-shot, since few-shot exemplars test inductive reasoning, not deduction/abduction~\cite{liuincomplete} in \fwd/\bwd.
Exemplars also do not improve \mm{} \bwd accuracy (Appendix~\ref{appendix:prompt_engineer}).
Prompts follow the same template as Figure~\ref{fig:intro} with format rules to parse outputs~\cite{liu2024we}.
% , and we let an \mm{} abstain (say ``IDK'') if it cannot give a valid question/answer.
%
Two \abr{nlp} graduate students write the prompts, with all design steps in Appendix~\ref{appendix:prompting}. 

\subsection{Evaluation Metrics} \label{subsection:metrics}

%We assess that \fwd and \bwd succeed if $a_{true} = a_{fwd}$ and $q_{bwd}$ is correctly answered by $a_{true}$, respectively. \ar{graduate students?}
To compute \fwd accuracy, two graduate students 
% researching \fwd 
annotate if 1280 \mm{} \fwd answers~$\hat{a}$ for a question~$q$ match its true answer $a$ (20 per answer type/model).\footnote{The 1280 total annotations are derived from 16 \mm{}s, 4 splits, and 20 annotations on each \mm{}/split combination.}
We test seven metrics~\cite{li2024pedantspreciseevaluationsdiverse} that evaluate if $\hat{a}$ and $a$ are equivalent.
We select DSPy-optimized~\cite{khattab2023dspy} GPT-4o for easy/hard entities and a rule-based method for numerical entities, since these methods had the highest agreement with humans (94\% on average).

For \bwd accuracy, students annotate if the answer to 1280 questions $\hat{q}$ from \bwd is $a$ (20 per answer type/model), following rules from \citet{li2024pedantspreciseevaluationsdiverse}.
% \jbgcomment{Did you use Zongxia's rules?  if so, cite that, as it will make it sound more rigorous}
We use DSPy-optimized GPT-4o as an oracle ($\textsc{Verify}^*(\hat{q}, a)$) to assess if $a$ answers~$\hat{q}$, which has high ($90\%$) human agreement.
Metric agreement is high but imperfect, so we also~show \fwd/\bwd~accuracy using our 1280 annotations in Figure~\ref{fig:benchmark_human}, which has the same trend as our metrics.

\input{data/consistency}

%% file: data/benchmark.tex
\begin{figure*}
    \centering
    \includegraphics[width=\linewidth]{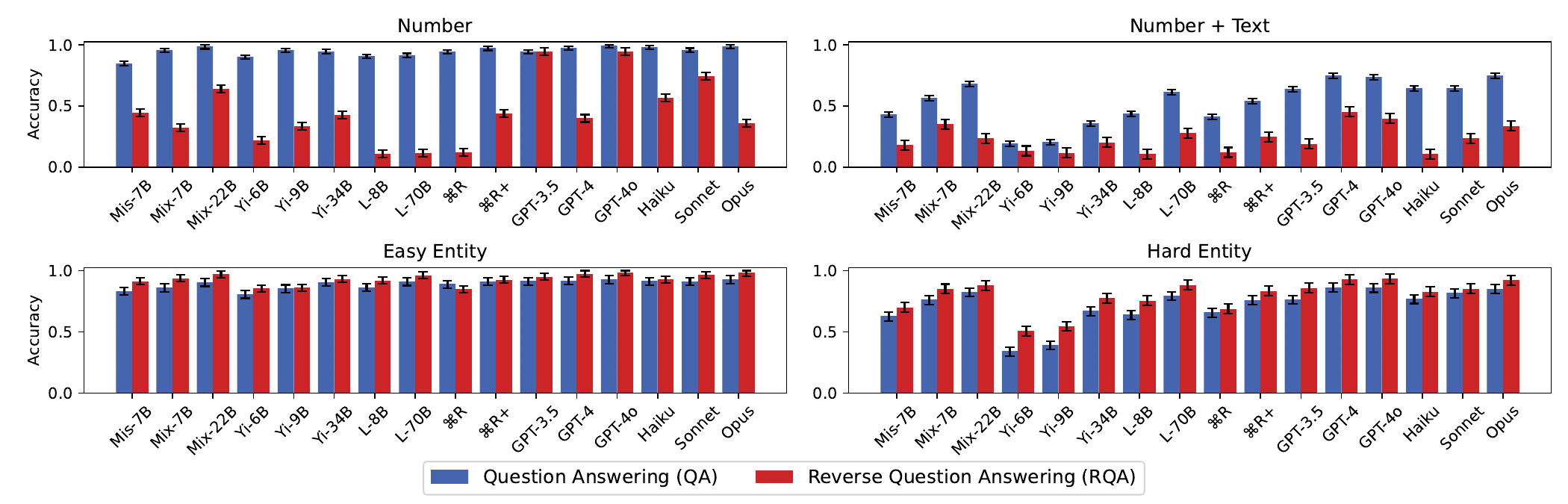}
    \vspace{-4ex}
        \caption{\small LLM \bwd (\textbf{\textcolor{blue}{blue}}) and \fwd (\textbf{\textcolor{red}{red}}) accuracy with 95\% CIs for metric error rate. LLMs are much weaker in \textit{abductive} \bwd in numerical settings (Number/Number+Text), but in text settings (Easy/Hard Entity), \textit{deductive} QA is slightly weaker.}
        % \jbgcomment{Is this QG AQ by another name?  If not, explain difference}
    \label{fig:benchmark}
    %\vspace{-2.5ex}
\end{figure*}

%% file: data/consistency.tex
\begin{figure*}
    \centering
    \includegraphics[width=\linewidth]{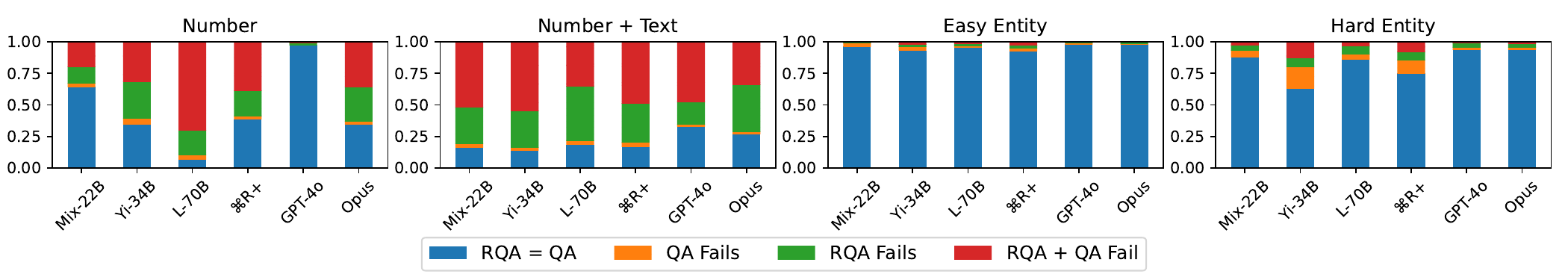}
    \vspace{-4ex}
        \caption{\small Logical consistency of \bwd and \fwd. For Number and Number+Text entities, most \mm{}s lack consistency (except GPT-4o), with \bwd as the main failure point. Otherwise, \mm{}s are fairly consistent, with \fwd as the failure point for Hard Factual Entities. We display the strongest \mm{} from each model family for brevity, with all results shown in Appendix~\ref{appendix:consistency}.}
    \label{fig:consistency}
    %\vspace{-1ex}
\end{figure*}

%% file: 2025_naacl_qgqa/sections/40_results.tex
\section{Evaluation of \fwd and \bwd} \label{section:results}

% \jbgcomment{It's usually a waste to have subsection immediately after section, can often be cut.  For ease of skimming, I'd replace the questions with the answers to the questions}

% \jbgcomment{For section titles, it would be good to have things that are more unique (e.g., "results" can be in just about any paper)}

Having designed our tasks (\cref{section:experimental_setup}), this
section tests \mm{}s abilities in \fwd and \bwd.
\mm{}s struggle in \bwd on numerical entities
(\cref{subsection:accuracy}) but surprisingly can often detect their
own errors (\cref{subsection:consistency}).
We study the types of entities that lead to \bwd errors
(\cref{subsection:correlators}) and qualitatively analyze differences between accurate and inaccurate questions from \bwd
(\cref{subsection:error_analysis}).

\subsection{\mm{}s Struggle with Numerical \bwd}\label{subsection:accuracy}

% \jbgcomment{Some of the paragraphs in this section don't have clear topic sentences and are describing the process more than the results.  The reader cares about the latter.}

We first see if \bwd ({\textcolor{red}{\textbf{red}}}, no stripe) or
\fwd ({\textcolor{blue}{\textbf{blue}}}, striped) is consistently
harder for \mm{}s (Figure~\ref{fig:benchmark}).
In numerical \settingName (Number, Number+Text), \mm{}s are much more
accurate in \fwd versus \bwd, revealing a clear abduction weakness.
Interestingly, in text \settingName (Easy, Hard), the trend
reverses---\bwd slightly beats \fwd 31/32 times.  Thus, \mm{}s cannot
be categorized as always stronger in generation or
validation~\cite{west2023generative, li2024benchmarking}: their
abilities are domain-specific.
If users (e.g. teachers) want to write question-answer pairs with
\mm{}s, we advise manually writing questions for numerical pairs and
answers for text pairs, and using \mm{}s to generate the counterparts,
given their strengths in numerical \fwd and textual~\bwd.

% \jbgcomment{If we have the room, it might be good to give examples of
%   the questions and answers they might use}

% This reflects the Generative \abr{ai} Paradox~\cite{west2023generative}, the idea that \mm{}s acquire generation (QG) better than understanding (\abr{qa}).
% The paradox does not always hold, so we believe it mainly applies in text, not numerical reasoning.
% Knowing \bwd's weakness in numbers and \fwd's in text can help practitioners decide how to allocate efforts to improve \mm{} reasoning reliability in their pipelines.
% In all, when \mm{} system designers can pick either abduction/deduction, we advise deduction for numbers and abduction~otherwise.\ar{What are scenarios where system designers are picking between abduction//deduction? This argument wasn't clear to me}

The Numbers domain has the largest \fwd/\bwd accuracy gaps, over 0.8
for LLaMA and Command-R.
Some view \mm{}s as strong math reasoners, but they excel just in
deductive \fwd tasks, as \fwd is the main testbed for math
abilities~\cite{ahn2024large}.
In contrast, abduction in textual \settingName appears in
instruction-tuning datasets with queries like ``Tell me about
Germany''.
Thus, researchers should design more \textit{abductive} math benchmarks, like \bwd, to holistically evaluate \mm{} math capabilities.
% Thus, \mm{}s overfit to math \textit{deduction}, failing to generalize to math \textit{abduction}.\ar{We can remove the previous sentence}
% In contrast, abduction with text entities is likely to appear in fine-tuning data with queries like ``Tell me about Germany.''
% Thus, we advise collecting diverse reasoning types for \mm{} training, as models cannot fully generalize across them.

\subsection{\fwd Can Self-Verify Numerical \bwd}
\label{subsection:consistency}

We chain \bwd and \fwd for consistency, i.e., see if
$\textsc{\fwd}(\bwd(a)) \approx a$
(Figure~\ref{fig:intro}).
If the check fails, the \bwd question $\hat{q}$
is invalid, the \mm{} fails to answer its own valid $\hat{q}$, or both failures occur.
We discuss how we disentangle these cases below.

%
% Recall for a given \mm{}, $\abr{qa}_{\mm{}}(q)=\hat{a}$ predicts an answer $\hat{a}$ given question $q$, and $AQ_{\mm{}}(a)=\hat{q}$ predicts, inversely, a question given an answer. Here we test a model's logical consistency by asking, given answer $a$, does $\abr{qa}_{\mm{}}(AQ_{\mm{}}(a))=a$? If these items do not match, then we must conclude that either the predicted question $\hat{q}$ is incorrect, or the model is unable to answer its own question.

\mm{}s give: 1) a question $\hat{q}$ with answer $a$; and 2) an answer $\hat{a}$ to their own $\hat{q}$ without using $a$.
We find three yes/no judgments via our metrics
(\cref{subsection:metrics}): a) does $a$ answer $\hat{q}$ (\bwd
succeeds); b) does $\hat{a}$ answer $\hat{q}$ (\fwd succeeds);
and c) are $a$ and $\hat{a}$ equivalent?
Answers $\mathcal{A} = (\text{a}, \text{b}, \text{c})$ to these judgments form a truth table to diagnose \mm{} inconsistencies, which in 91\% of cases, fall into the four cases of $\mathcal{A}$ below:

\begin{enumerate}
    \item $(\texttt{y}, \texttt{y}, \texttt{y})$: \bwd = \fwd (consistent).
    \item $(\texttt{n}, \texttt{y}, \texttt{n})$: Just \bwd fails.
    \item $(\texttt{y}, \texttt{n}, \texttt{n})$: Just \fwd fails.
    \item $(\texttt{n}, \texttt{n}, \texttt{n})$: \bwd \textit{and} \fwd fail.
\end{enumerate}

Other rare cases of $\mathcal{A}$ are metric prediction errors or errors in $\hat{q}$ (e.g. ambiguity), which we omit for this analysis. Appendix~\ref{appendix:consistency} shows all cases of $\mathcal{A}$.

\mm{}s are fairly consistent in textual \settingName, but often fail
the check in numerical \settingName, except GPT-4o
(Figure~\ref{fig:consistency}, left).
Thus, our \mm{}s are logically inconsistent in numerical abduction and
deduction.
In such cases, \fwd rarely fails alone:~either both \bwd and \fwd
fail, where the \mm{} gives an invalid question that it cannot answer,
or just \bwd fails, where the \mm{} detects its error.
The latter is akin to hallucination
snowballing~\cite{zhang2023language}---inaccurate questions are not
just due to knowledge gaps, as \mm{} can answer their invalid
question accurately (e.g. 33\% of cases for Opus).

For instance, given the answer ``127 countries'', Opus incorrectly produces the question ``How many countries are members of the United Nations that do not have veto power in the UN Security Council?''. However, when Opus answers its own question, it knows there are 193 countries in the UN and five of them have veto power,\footnote{At the time of writing this paper.} returning the correct answer of ``188''.
Thus, self-verification~\cite{weng-etal-2023-large} could be a useful way to verify the correctness of responses in numerical
\bwd tasks.

%Thus, self-verification~\cite{weng-etal-2023-large} could boost numerical QG accuracy.
%Our results reinforce \cref{subsection:accuracy}; numerical \bwd can be so unreliable that \mm{}s detect their own question inaccuracies.

\input{data/correlators}

\subsection{Number+Text \bwd Errs on Rare Entities} \label{subsection:correlators}

To find when \mm{}s fail in numerical \bwd~(\cref{subsection:accuracy}), we test two indicators of \bwd error.
We first see how often $a$ appears in the Dolma pretraining corpus~\cite{soldaini-etal-2024-dolma} via infini-gram~\cite{liu2024infini}, a proxy for the size of all valid questions an \mm{} must abductively reason over in \bwd.
Next, as \cref{subsection:consistency} hints \mm{}s may give overly-hard questions (RQA+QA fail), we use the Prometheus \mm{}~\cite{kim2024prometheus} to get a 1-5 difficulty score for $\hat{q}$.
We average metrics pivoted by \bwd~success/failure on the subset containing human annotations (\cref{subsection:metrics}).

Number+Text $a$ have lower Dolma token counts when \bwd fails (Fig~\ref{fig:correlators}), so \mm{}s struggle to recall long-tail numerical facts~\cite{kandpal2023large}.
In Numbers, \bwd $\hat{q}$ are harder when \bwd fails.
Thus, calibrating \mm{}s with desired difficulty~\cite{srivastava-goodman-2021-question} could help designers avoid errors from overly-hard questions in \bwd on numbers. 
Also, difficulty and token count are similar in \bwd success/failure for Numbers+Text and Numbers, respectively, so \bwd errors depend on answer type, like in \fwd~\cite{vakulenko2020wrong}.

\subsection{\mm{}s Fail to Write Multi-Step Questions} \label{subsection:error_analysis}

% \jbgcomment{The way this subsection is titled, it feels odd to to contrast with question types in \abr{qa} (Rogers has a taxonomy)}

% \jbgcomment{I think this could use more meat.  I realize the space is tight, but even writing the "real" version in the appendix (with lots of examples) could lead to a better "real" version here.  At the moment it's basically a recounting of "what" you found, but not enough "why".  I think if we can have a better recounting of why this stuff happens and how this can improve training would make this much stronger.}

% \jbgcomment{I'd also like to see a little more discussion of diversity.  Which of the AQ examples are near matches for something in the training data, how often does a model generate something different from other models, how often are there hallucinations / ambiguities / false presup / etc. }

For qualitative insights into question types $\hat{q}$ from \bwd, we analyze 30 $\hat{q}$ when \bwd fails/succeeds in strong \mm{}s with low \bwd accuracy (\cref{subsection:accuracy}): L-70B, GPT-4, and Opus.
For brevity, we just study the Numbers split, as its similar answers yield $\hat{q}$ with similar patterns, and group $\hat{q}$ as: 1) \textbf{Single-Step}: has one math operation; 2) \textbf{Multi-Step}: has 2+ math~operations; 3) \textbf{Fact-Based}: tests factual knowledge; and 4) \textbf{Metric Error}: metric misclassification.
In Appendix~\ref{appendix:qual}, we analyze more questions $\hat{q}$ in aspects like question novelty, answerability, similarity across models, and memorization.

\input{data/errors}

When \bwd fails, $\hat{q}$ is often multi-step (Fig~\ref{fig:errors})---combining math and facts (\textit{how many legs are on a human, cat, \& spider?}) or adding primes~(Fig~\ref{fig:intro}).
In contrast, valid $\hat{q}$ are often single-step (\textit{what is $19^2$?}) or factual~\cite{mccarthy1959programs} (\textit{how many days is a leap year?} for \emph{366}).
We believe the errors in multi-step \bwd are from preference tuning; users favor a complex output even if it is wrong~\cite{wen2024languagemodelslearnmislead}.
Thus, curbing complexity bias in alignment, or multi-hop \fwd decoding methods~\cite{zhao-etal-2021-multi-step}, may improve \mm{}s in multi-step \bwd.

%% file: data/correlators.tex
\begin{figure}
    \vspace{-1ex}
    \centering
\includegraphics[width=\linewidth]{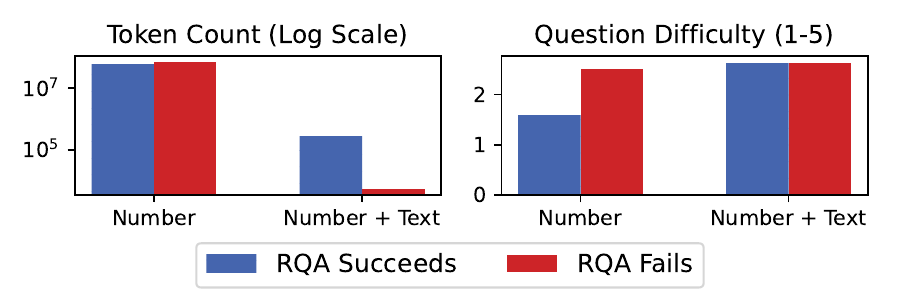}
    % \jbgcomment{be very careful of vspace, better to edit PDF or includegraphics call to be on safe side}
    \vspace{-4.75ex}
    % \jbgcomment{Token count of what?}
    \caption{\small Answer answer token count in Dolma and question difficulty of when \bwd succeeds/fails, averaged over \mm{}s.}
    \label{fig:correlators}
    \vspace{-1ex}
\end{figure}

%% file: data/errors.tex
\begin{figure}
    \centering
    \includegraphics[width=\linewidth]{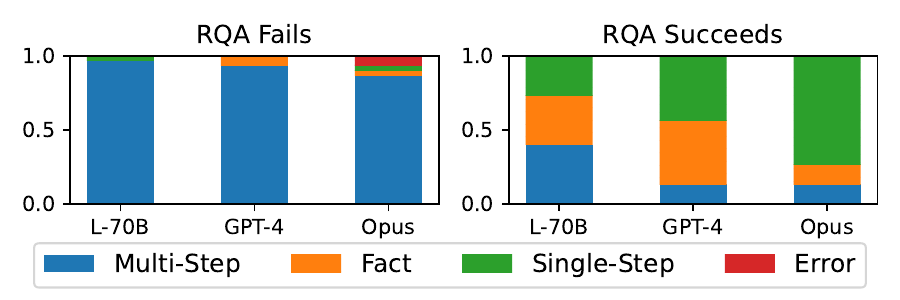}
    \vspace{-4.75ex}
    \caption{\small Analysis of Number \bwd errors. \bwd often fails when the \mm{} tries to give a complex, multi-step question.}
    % \jbgcomment{Need a takeaway and describing the categories.  figures should be skimmable by a reader who hasn't looked at the rest of the paper.}
    \label{fig:errors}
    %\vspace{-1ex}
\end{figure}

%% file: 2025_naacl_qgqa/sections/50_related_work.tex
\section{Related Work}

\textbf{\mm{} Reasoning:} Several works have explored \mm{} reasoning to improve accuracy~\cite{qiao-etal-2023-reasoning} or explainability~\cite{si-etal-2024-large}.
More recently, works explore if \mm{}s can execute diverse reasoning strategies, including inductive~\cite{bowen-etal-2024-comprehensive, yang-etal-2024-language}, deductive~\cite{sanyal-etal-2022-fairr, mondorf-plank-2024-comparing}, and abductive~\cite{zhao-etal-2023-abductive, balepur-etal-2024-artifacts} reasoning.
However, we are the first to pinpoint abduction abilities via \bwd, which differs from traditional question generation setups as we do not have access to an input context~\cite{zhang2021review}.\\

\noindent \textbf{\mm{} Consistency:} \mm{}s must be consistent~to reliably help users~\cite{visani2022statistical}, but \mm{}s are inconsistent under perturbations like prompt format~\cite{sclar2024quantifying}, entity reversal~\cite{berglund2024the}, negation~\cite{ravichander-etal-2022-condaqa, balepur-etal-2024-easy}, and ordering~\cite{zheng2024large}.
Recent work finds inconsistencies~in \mm{} generation and verification in math, \fwd, style transfer, and coding~\cite{li2024benchmarking, pmlr-v235-gu24c}, which we reproduce via an \bwd/\fwd consistency check.
\citet{deb2023fill} and~\citet{yu2024metamath} similarly compare \mm{}s in forwards (\fwd) and backwards (filling question blanks for an answer) reasoning in math.
While~\citet{deb2023fill} claim backwards reasoning is abductive, we argue it is deductive as there is just one answer; we more aptly test abduction/deduction consistency via \bwd/\fwd.

% \jbgcomment{Even if it doesn't go into the main paper, it would be good to have coverage in appendix of how questions can be bad: false presuppositions, ambiguity, etc.}

% \jbgcomment{I'd also talk about question generation}

%% file: 2025_naacl_qgqa/sections/60_conclusion.tex
\section{Conclusion}

We test \mm{} \bwd and \fwd abilities.
\mm{}s have notably low accuracy in numerical \bwd~which is not just due to knowledge gaps, as models can often answer their own invalid questions correctly.
These weaknesses can be excised in future benchmarks to more holistically evaluate \mm{} numerical abductive reasoning and math capabilities.
%Thus, \mm{} reasoning is not yet fully generalizable; \mm{}s need specific training on numerical abduction, as they fail to generalize from deduction tasks alone.
To reduce inaccuracies in numerical \bwd, often from generating overly-complex questions, we suggest calibrating models using difficulty scores, collecting user preferences that control for complexity bias, and adapting prior multi-hop \abr{qa} methods---key steps for reliable \mm{} reasoning in downstream~tasks.

%% file: 2025_naacl_qgqa/sections/70_limitations_ethics.tex
\section{Limitations}

\mm{}s are sensitive to prompt formats~\cite{sclar2023quantifying}, so varying prompts could impact \mm{} accuracy in \bwd and \fwd.
To ensure our prompts are reliable, we followed best practices~\cite{schulhoff2024prompt} and kept refining prompts as \mm{} errors surfaced; the full prompt engineering process is documented in Appendix~\ref{appendix:prompting}.
Our final prompts will be released and are considered very reasonable implementations of \bwd and \fwd.
Further, in Appendix~\ref{appendix:prompt_engineer}, we test if common prompt engineering strategies (few-shot exemplars, chain-of-thought) can alleviate the low numerical \bwd accuracy of GPT-4 but find minimal benefits, suggesting that accuracy gaps between \fwd and \bwd cannot be attributed to prompt formatting alone.

% Further, we acknowledge that several analyses could be conducted beyond question accuracy in \bwd, but we are limited by space constraints; we encourage this exploration for future research.
% We felt that the contributions of our work---revealing the first signs of accuracy gaps and inconsistencies in \bwd and \fwd along with preliminary analysis as to when/why \bwd fails---aligns well with ACL guidelines of a short paper: a small, focused contribution and an interesting application nugget.\footnote{\url{https://aclrollingreview.org/cfp\#short-papers}}
% We also provide more extensive qualitative evaluations in Appendix~\ref{appendix:qual} on question answerability, cross-model question duplicates, and question novelty, which we would be happy to include in the final version with the extra space.

\section{Ethical Considerations}

\bwd uses abduction, a core reasoning strategy that aims to arrive at a plausible explanation given a set of facts.
However, our current findings suggest that \mm{} abductive reasoning in numerical settings is highly unreliable.
We advise practitioners~to take caution when using \mm{}s to reason via numerical abduction in downstream tasks, including designing math exam questions, explaining financial forecasts, proposing economic policies, or diagnosing medical patients from numerical data.

\section*{Acknowledgements}

We would like to thank the \abr{clip} lab at University of Maryland and our external collaborators for their feedback, including Neha Srikanth, Haozhe An, Yu Hou,  Abhilasha Sancheti, Connor Baumler, Seraphina Goldfarb-Tarrant, Aidan Peppin, Jack Wang, and Vishakh Padmakumar.
This material is based upon work supported by the National Science Foundation under Grant No. \abr{iis}-2403436 (Boyd-Graber), \abr{iis}-2339746 (Rudinger), \abr{dms}-2134012 (Ravichander), and \abr{dge}-2236417 (Balepur).
Any opinions, findings, and conclusions or recommendations expressed
in this material are those of the author(s) and do not necessarily
reflect the views of the National Science Foundation.
Cloud computing resources were made possible by a gift from Adobe Research.
Access to Command-R was made possible with a Cohere for AI Research Grant.

%% file: 2025_naacl_qgqa/sections/80_appendix.tex
\section{Appendix}

\subsection{Dataset Details} \label{appendix:dataset}

We show details for our dataset in Table~\ref{table:dataset}.
Our entities are derived from Quizbowl questions~\cite{Rodriguez2019QuizbowlTC} from the \qb{} Reader \abr{api}\footnote{\url{https://www.qbreader.org/api-docs/}}, which is free to use and publicly available online.
We verify that all questions are answerable by the given answer via Google search.
If any question was found to be unanswerable, we manually edited the question such that it was answerable.
Thus, all of our collected data is within their license and terms of use, and our use of these questions are within their intended use.
Since expert trivia writers curated these questions for academic competitions, we did not need to check that our data has PII.
All questions and answers are in English.

\subsection{Prompting Details} \label{appendix:prompting}

Below, we document our prompt engineering process for the \fwd and \bwd prompts shown in Figure~\ref{fig:intro}.
To assess each prompt version, we ran inference on a small subset of examples with the Yi and LLaMA \mm{}s and manually assessed the quality of questions/answers to identify prevalent issues that could be avoided through prompt engineering.
In all adjacent prompt boxes below, \textbf{\textcolor{blue}{blue text}} corresponds to us adding instructions to the previous version of the prompt, and \textbf{\sout{\textcolor{red}{red text}}} corresponds to us removing instructions from the previous version.

Our initial \bwd prompt is in Prompt~\ref{prompt:qg_one}.
With this prompt, our \mm{}s generated verbose answers, so we added the instruction that all questions must be ``one-sentence'' (Prompt~\ref{prompt:qg_two}).
Next, we observed that it was difficult to reliably parse the question from the model's generated output, so we added formatting constraints (Prompt~\ref{prompt:qg_three}).
At this point, when we looked at the model's generated questions more closely, we saw that models could cheat---adding the answer in the question itself (e.g. giving the question ``How many of the 150 people attended the conference'' for the answer ``150 people'').
Thus, we added an instruction to forbid this behavior (Prompt~\ref{prompt:qg_four}).
Finally, as we noticed many of the questions were inaccurate, we wanted to study if abstention could alleviate these issues, so we added an instruction (Prompt~\ref{prompt:qg_five}) allowing the model to respond with ``IDK'' \cref{subsection:models}.
We added abstention to test \mm{} calibration~\cite{feng-etal-2024-dont}, but abstention rates are only 3\% in \abr{qa} and <1\% in \bwd, so we do not study it in this work.
We keep abstention to avoid re-running all \mm{}s and omit rare cases of abstention.
Our final \bwd prompt is in Prompt~\ref{prompt:qg}.

We then designed our \fwd prompt by mimicking the format of the final \bwd prompt, shown in Prompt~\ref{prompt:qa_one}.
We initially wrote the constraint that the answer must be ``short'' and ``just a few words,'' but we felt these instructions were ambiguous, and the easy and hard entities split of our dataset had answers that were longer than just a few words; as a result, we removed these instructions, and used ``the'' instead of ``a'' to make it clear that there is only one valid answer (Prompt~\ref{prompt:qa_two}).
After removing these instructions, we noticed that models would often generate very long explanations before or after answering the question.
To avoid this, we added an instruction stating that we were just looking for the answer and no explanation (Prompt~\ref{prompt:qa_two}).
Our final \fwd prompt is in Prompt~\ref{prompt:qa}.

\subsection{Model Details} \label{appendix:experimental_setup}

The \mm{}s used in this work are from the following endpoints:
\begin{itemize*}
  \item LLaMA-8B: \texttt{Meta-Llama-3-8B-Instruct}
  \item LLaMA-70B: \texttt{Meta-Llama-3-70B-Instruct}
  \item Mistral-7B:\\\texttt{Mistral-7B-Instruct-v0.3}
  \item Mixtral-8x7B: \texttt{Mixtral-8x7B-Instruct-v0.1}
  \item Mixtral-8x22B: \texttt{Mixtral-8x22B-Instruct-v0.1}
  \item Yi-6B: \texttt{Yi-1.5-6B-Chat}
  \item Yi-9B: \texttt{Yi-1.5-9B-Chat}
  \item Yi-34B: \texttt{Yi-1.5-34B-Chat}
  \item Command-R: \texttt{command-r}
  \item Command-R+: \texttt{command-r-plus}
  \item GPT-3.5: \texttt{gpt-3.5-turbo-0125}
  \item GPT-4: \texttt{gpt-4-turbo-2024-04-09}
  \item GPT-4o: \texttt{gpt-4o-2024-05-13}
  \item Haiku: \texttt{claude-3-haiku-20240307}
  \item Sonnet: \texttt{claude-3-sonnet-20240229}
  \item Opus: \texttt{claude-3-opus-20240229}
\end{itemize*}

LLaMA, Mistral, and Yi models are accessed via huggingface, and all other models are accessed through their respective \abr{api} endpoints.
We allocated 8 NVIDIA:A6000s for Mixtral-8x22B, 8 NVIDIA:A5000s for Mixtral-8x7B, Yi-34B, and LLaMA-70B, 2 NVIDIA:A6000s for Yi-9B and LLaMA-8B, and 1 NVIDIA:A6000 for all other non-\abr{api} models (which were run on CPU only).
Each model was allocated 24 hours to run both \fwd and \bwd on our dataset.

\mm{}s generate with 0 temperature, a minimum token length of 5, and a maximum token length of 5. All other unspecified parameters are set to their respective default values.

\subsection{Metric Details} \label{appendix:metrics}

To design a metric for \fwd accuracy, we consider seven answer equivalence metrics, which check if a candidate answer $a_{cand}$ is semantically equivalent to a ground-truth answer $a_{true}$: 1) DSPy-optimized GPT-4o; 2) A rule-based method designed specifically for each dataset; 3) Exact match; 4) Token F1 score; 5) Token Recall Score; 6) Token Precision Score; and 7) \textsc{PEDANTS}~\cite{li2024pedantspreciseevaluationsdiverse}, a classifier designed for answer equivalence.
The DSPy method in (1) uses a maximum of 10 bootstrapped demos, a maximum of 10 labeled demos, and 20 candidate programs; it uses 64 examples for training (seeding the prompts) and 64 examples for validation.
We decide the optimal decision thresholds for (4), (5), and (6) using the 64 validation examples.
We present the agreement with human annotations of each metric in Table~\ref{table:metrics}, which is how we picked the metric to use for each dataset split.
In all, our \fwd accuracy metric has 94\% raw agreement with humans on 1152 held-out examples.

Since there are no automated metrics to check whether a question $q$ can correctly be answered by an entity $a$, we design our own metric for \bwd accuracy.
Given the strength of the DSPy GPT-4o approach in \fwd accuracy, we similarly design a DSPy-optimized GPT-4o classifier that determines if $q$ is correctly answered by $a$, using the same hyperparameters for \fwd accuracy.
Overall, this \bwd accuracy metric has 90\% raw agreement with humans on 1152 held-out examples.
We also considered Jury approaches~\cite{Verga2024ReplacingJW}, which ensemble multiple \mm{}s instead of relying just on a single \mm{}.
However, using majority vote with three/five \mm{}s boosted our metric's accuracy by less than 2\%, which we did not feel justified the much larger computational expenses.

All metrics are reported for a single run, and we provide confidence intervals in Figure~\ref{fig:benchmark} corresponding to the error rates in our metrics.

\subsection{Abduction/Deduction Human Accuracy} \label{appendix:human_accuracy}

In Figure~\ref{fig:benchmark_human}, we show a version of Figure~\ref{fig:benchmark} using our human annotations on a subset of data versus the automated metrics on the entire splits.
Our trend holds on the human-annotated subset; \mm{}s are still much weaker in numerical \bwd versus \fwd, but their \fwd capabilities slightly beat \bwd in the text-based settings.

\subsection{\bwd with Prompting Engineering} \label{appendix:prompt_engineer}

To explore if \bwd weaknesses can be alleviated with prompt engineering efforts~\cite{schulhoff2024prompt}, we test three prompting strategies: 1) Zero-Shot Chain-of-Thought Prompting (asking the \mm{} to ``Think step by step'' before answering); 2) Self-Verification (asking the \mm{} to ``Check if the question is accurate after generating a question''); and 3) Five-Shot Prompting (including five exemplars showing the model how to generate a question for an answer).
To write exemplars for (3), we pick question/answer pairs when $\bwd$ succeeds in the zero-shot setting to make the priors in the exemplars most similar to the model's original generations.
The prompts for (1), (2), and (3) are in Prompts~\ref{prompt:qg_cot}, \ref{prompt:qg_verify}, and \ref{prompt:qg_fewshot}, respectively.

We experiment with GPT-4 on Numbers and Numbers+Text, as the model showed a surprising \bwd weakness in these settings. GPT-4 is also considered to respond well to prompt engineering efforts, making it a suitable candidate for our prompting strategies.
Overall, none of these prompting strategies can close the accuracy gap between \bwd and \fwd (Figure~\ref{fig:benchmark_prompt}).
Chain-of-thought prompting increases GPT-4's \bwd accuracy by $\sim0.15$, but it is still significantly lower than \fwd, which does not use chain-of-thought.
This shows that the accuracy gap between \fwd and \bwd may be an inherent reasoning flaw of current \mm{}s that cannot be fully mitigated via prompt engineering.

\subsection{Full Consistency Analysis} \label{appendix:consistency}

In this section, we describe the consistency analysis for all values of our truth table $\mathcal{A}$, introduced in \cref{subsection:consistency}.
Apart from the four categories described before, the truth table outcome can also be ``Ambiguous Question'' if $\mathcal{A} = (\texttt{y}, \texttt{y}, \texttt{n})$, as both steps succeeded but converged to different answers (meaning the question had more than one possible correct answer).
Another option is for the mistakes to cancel out, which is a rare scenario $\mathcal{A} = (\texttt{n}, \texttt{n}, \texttt{y})$ where the model generated an inaccurate question and answered its own question incorrectly, but managed to arrive at the original entity $a$. The final category is a Metric Prediction Error, a scenario that only occurs if either just \fwd or \bwd was predicted to fail, but $a$ and $\hat{a}$ were predicted to be matching ($\mathcal{A} = (\texttt{n}, \texttt{y}, \texttt{y})$ or $\mathcal{A} = (\texttt{y}, \texttt{n}, \texttt{y})$). These scenarios are summarized in Table~\ref{table:truth_table}.

Figure~\ref{fig:consistency_all} reports the full consistency analysis for all 16 of our \mm{}s and all truth table scenarios.
The four categories reported in Figure~\ref{fig:consistency} encompass most of the truth table.
Further, even for smaller \mm{}s, our claims hold; \mm{}s can often detect their own question inaccuracies from \bwd through \fwd.

\subsection{Further Analysis of \bwd Questions} \label{appendix:qual}

Due to page limit constraints of a short paper, we were unable to show the entire qualitative analysis we conducted on questions generated in \bwd.
Below, we give more qualitative results on the answerability of questions from \bwd (Appendix~\ref{appendix:unanswerable}), a cross-model comparison of question duplicates in \bwd (Appendix~\ref{appendix:duplicate}), the ability of \mm{}s to match the ground-truth question during \bwd (Appendix~\ref{appendix:match_gold}), and a brief investigation into memorization in the \bwd task (Appendix~\ref{appendix:memorize}).

\subsubsection{Are \bwd questions unanswerable?} \label{appendix:unanswerable}

We now seek to understand the types of \bwd questions generated in the Number+Text setting, complementing our analysis in \cref{subsection:error_analysis}.
The Number+Text questions have higher variance and cannot be as neatly categorized as in \cref{subsection:error_analysis} (e.g. single-step computation). So instead, we study the \textit{answerability} of 30 generated questions from each \mm{}, i.e., if the question is clear but leads to an incorrect answer, or if the question has an issue that makes it difficult to answer.
We adopt five categories of unanswerable questions from~\citet{rogers2023qa}:\\
\noindent 1) \textbf{Invalid Premise:} the question contains a false assumption, so it is impossible to answer. For example, Opus generates the question \textit{How old was the world's oldest tortoise, Jonathan, when he passed away in 2022?}, but this Tortoise is still alive. \\
\noindent 2) \textbf{No Consensus on the Answer:} the question does not have a single, agreed-upon answer . For example, LLaMA generates the question \textit{What is the unique property of the Lie algebra E8 that makes it particularly interesting in theoretical physics?}, but Lie algebra has many distinct, interesting properties that would answer the question. \\
\noindent 3) \textbf{Information not yet Discovered:} the answer to this question is not yet known. For example, GPT-4 generates the question \textit{How long, in terms of word count, is the sentence that holds the record for being the longest in the English language without using any punctuation?}, but it is not yet known what could theoretically be the longest sentence. \\
\noindent 4) \textbf{Missing Information:} the question does not have enough information, or it is too vague. For example, GPT-4 generates the question \textit{How many individuals attended the annual community festival last year according to the final headcount?}, which cannot be answered without knowing more details.\\
\noindent 5) \textbf{Answerable:} The question has one right answer.

As expected, when \bwd succeeds, questions are mostly answerable (Figure~\ref{fig:errors_numtext}).
However, a non-trivial proportion of generated questions when \bwd fails are unanswerable, reaching nearly 60\% for GPT-4.
The most common types of unanswerable questions are those that are missing information, meaning that they are too vague or ambiguous, or those that have false premises or assumptions. 
While several works explore methods to \textit{answer} ambiguous questions~\cite{min-etal-2020-ambigqa, kim-etal-2023-tree} or questions with false presuppositions~\cite{yu-etal-2023-crepe, kim-etal-2023-qa}, our analysis reveals a need to \textit{avoid generating} ambiguous or faulty-presupposition questions in \bwd.

In Tables~\ref{table:error_num_qual} and \ref{table:error_num_text_qual}, we provide examples of question/error types in our qualitative analysis on the Number and Number+Text split, respectively.

\subsubsection{Do \mm{}s give the same \bwd questions?} \label{appendix:duplicate}

While most of our analysis treated \mm{}s independently, we now study whether \mm{}s generate the same exact questions (i.e. duplicates) in \bwd.
Figure~\ref{fig:overlap} shows that \mm{}s more frequently generate duplicated questions across entities versus matching questions from other models. For example, LLaMA-3 70B generates 379 duplicate questions in the Numbers setting, even when the input answer is altered. This aligns with very recent work suggesting that \mm{}s may often conduct pattern-matching rather than engaging in true, generalizable reasoning~\cite{mirzadeh2024gsm}.

Interestingly, models in the same family are more likely to generate duplicated questions. For example, GPT-3.5, GPT-4, and GPT-4o generate the same questions in \bwd more often than when compared to other \mm{} families.
Thus, we speculate that these model families likely share similar pre-training and alignment data, which is optimized on through different training recipes.

\subsubsection{Does \bwd match the gold question?} \label{appendix:match_gold}

We now explore whether the questions generated for an answer in \bwd match the gold question we collected for that answer.
When determining if the two questions are semantically equivalent, we follow the protocol of~\citet{balepur-etal-2024-artifacts} and analyze whether the two questions test the exact same knowledge.
Figure~\ref{fig:errors_match} shows that the \mm{}s can often match the true question when \bwd succeeds in Number+Text settings, reaching as high as 40\% of cases for GPT-4; the questions never matched for Number.
One explanation for the high match rate is dataset contamination~\cite{ishihara-2023-training}, but it is also possible that the most likely question the \mm{} abductively reason towards is the ground-truth question.
For example, for the answer ``120 counties,'' the only salient fact linked to the entity is that Kentucky has 120 counties~\cite{mccarthy1959programs}; this led GPT-4's question and the ground-truth question to both ask about Kentucky.

\subsubsection{Are any \bwd questions memorized?} \label{appendix:memorize}

Since the duplicates in Appendix~\ref{appendix:match_gold} suggest that \mm{}s may just be retrieving similar questions from pretraining rather than reasoning towards new questions in \bwd, we now investigate the \textit{novelty} of the \bwd questions~\cite{merrill2024evaluating}, i.e., whether they are exactly copied from pretraining.
We do not know which corpora all of our \mm{}s are trained on, so we use the Dolma~\cite{soldaini-etal-2024-dolma} corpus as a proxy for pretraining data.
For each generated \bwd question $\hat{q}$, we compute how frequently the exact question $\hat{q}$ appears in Dolma via infini-gram~\cite{liu2024infini}.

Table~\ref{table:question_memorization} reveals in total, $2.87$\% of \bwd questions are exactly found in Dolma. For comparison, $1.25$\% of our ground-truth questions exist in Dolma.
While we did not explicitly prompt the model to give a new question that it has not seen in pretraining, practitioners may need to design specialized techniques if they desire novel \bwd questions.

When comparing exact question match frequency by model, weaker/smaller \mm{}s tend to copy more from pretraining data, suggesting that smaller \mm{}s are more prone to \bwd memorization.
Further, the Hard Fact setting is much less prone to question copying in \bwd, likely because the \bwd input answers have very low pretraining token count (\cref{subsection:correlators}), which further supports that \mm{}s may struggle to retrieve exact pretraining knowledge for long-tail facts~\cite{kandpal2023large}.

We present examples of \bwd questions that appear the most in Dolma in Table~\ref{table:memorization}.
The tendency to generate inaccurate or ambiguous questions may be influenced by pretraining, as many of these questions appear directly in Dolma.

\clearpage

\input{appendix/prompts}
\input{appendix/dataset}
\input{appendix/metric_eval}
\input{appendix/benchmark_human}
\input{appendix/prompt_eng}
\input{appendix/truth_table}
\input{appendix/consistency_all}

\input{appendix/errors_numtext}
\input{appendix/error_qual}
\input{appendix/match}
\input{appendix/overlap}
\input{appendix/question_mem}
\input{appendix/memorization_ex}

%% file: appendix/prompts.tex
\hypersetup{
    colorlinks=true, % Enable colored links
    linkcolor=white, % Default color for internal links (sections, etc.)
    citecolor=white, % Default color for citations
    urlcolor=white % Default color for external URLs
}

\begin{prompt}[title={Prompt \thetcbcounter: Reverse Question Answering Prompt V1 (\bwd)}, label=prompt:qg_one]
\texttt{Generate a question with the answer: ``$a$''.}
\end{prompt}

\begin{prompt}[title={Prompt \thetcbcounter: Reverse Question Answering Prompt V2 (\bwd)}, label=prompt:qg_two]
\texttt{Generate a \textbf{\textcolor{blue}{one-sentence}} question with the answer: ``$a$''.}
\end{prompt}

\begin{prompt}[title={Prompt \thetcbcounter: Reverse Question Answering Prompt V3 (\bwd)}, label=prompt:qg_three]
\texttt{Generate a one-sentence question with the answer: ``$a$''. \textbf{\textcolor{blue}{Please format your output as ``Question: [insert generated question]''}}}
\end{prompt}

\begin{prompt}[title={Prompt \thetcbcounter: Reverse Question Answering Prompt V4 (\bwd)}, label=prompt:qg_four]
\texttt{Generate a one-sentence question with the answer: ``$a$''. \textbf{\textcolor{blue}{The question should not contain the text ``$a$''.}} Please format your output as ``Question: [insert generated question]''}
\end{prompt}

\begin{prompt}[title={Prompt \thetcbcounter: Reverse Question Answering Prompt V5 (\bwd)}, label=prompt:qg_five]
\texttt{Generate a one-sentence question with the answer: ``$a$''. The only possible answer to the question must be ``$a$''. The question should not contain the text ``$a$''. Please format your output as ``Question: [insert generated question]''. \textbf{\textcolor{blue}{If no possible question exists say ``IDK''.}}}
\end{prompt}

\begin{prompt}[title={Prompt \thetcbcounter: Final Reverse Question Answering Prompt (\bwd)}, label=prompt:qg]
\texttt{Generate a one-sentence question with the answer: ``$a$''. The only possible answer to the question must be ``$a$''. The question should not contain the text ``$a$''. Please format your output as ``Question: [insert generated question]''. If no possible question exists say ``IDK''.}
\end{prompt}

\clearpage

\begin{prompt}[title={Prompt \thetcbcounter: Question Answering Prompt V1 (\fwd)}, label=prompt:qa_one]
\texttt{Generate a short answer to the question: ``$q$''. The answer should just be a few words long. Please format your output as ``Answer: [insert generated answer]''. If no possible answer exists say ``IDK''.}
\end{prompt}

\begin{prompt}[title={Prompt \thetcbcounter: Question Answering Prompt V2 (\fwd)}, label=prompt:qa_two]
\texttt{Generate \textbf{\textcolor{red}{\sout{a short}} \textbf{\textcolor{blue}{the}}} answer to the question: ``$q$''. \textbf{\textcolor{red}{\sout{The answer should just be a few words long.}}} Please format your output as ``Answer: [insert generated answer]''. If no possible answer exists say ``IDK''.}
\end{prompt}

\begin{prompt}[title={Prompt \thetcbcounter: Question Answering Prompt V3 (\fwd)}, label=prompt:qa_three]
\texttt{Generate the answer to the question: ``$q$''. \textbf{\textcolor{blue}{Give just the answer and no explanation.}} Please format your output as ``Answer: [insert generated answer]''. If no possible answer exists say ``IDK''.}
\end{prompt}

\begin{prompt}[title={Prompt \thetcbcounter: Final Question Answering Prompt (\fwd)}, label=prompt:qa]
\texttt{Generate the answer to the question: ``$q$''. Give just the answer and no explanation. Please format your output as ``Answer: [insert generated answer]''. If no possible answer exists say ``IDK''.}
\end{prompt}

\clearpage

\begin{prompt}[title={Prompt \thetcbcounter: \bwd with Chain-of-Thought}, label=prompt:qg_cot]
\texttt{Generate a one-sentence question with the answer: ``$a$''. The only possible answer to the question must be ``$a$''. The question should not contain the text ``$a$''. Think step by step and reason before generating the question. After reasoning, please format your final output as ``Question: [insert generated question]''.}
\end{prompt}

\begin{prompt}[title={Prompt \thetcbcounter: \bwd with Self-Verification}, label=prompt:qg_verify]
\texttt{Generate a one-sentence question with the answer: ``$a$''. The only possible answer to the question must be ``$a$''. The question should not contain the text ``$a$''. Please format your output as "Question: [insert generated question]". After generating a question, answer your own question to verify that the answer is ``$a$'', formatted as "Answer: [insert answer to generated question]".}
\end{prompt}

\begin{prompt}[title={Prompt \thetcbcounter: \bwd with Five Exemplars}, label=prompt:qg_fewshot]
\texttt{Generate a one-sentence question with the answer: ``$a$''. The only possible answer to the question must be ``$a$''. The question should not contain the text ``$a$''. Please format your output as ``Question: [insert generated question]''.}

\texttt{\\Answer: 328\\Question: What is the sum of the first 15 prime numbers?\\ \\ Answer: 710 survivors \\
Question: How many people survived the sinking of the RMS Titanic in 1912? \\ \\ Answer: 648\\
Question: What is the product of 12 and 54?\\ \\Answer: 286 ayats \\
Question: How many verses are there in the longest chapter of the Quran, Surah Al-Baqarah? \\ \\ Answer: 311
\\ Question: What is the sum of the first three prime numbers greater than 100? \\ \\ Answer: $a$ \\
Question:}

\end{prompt}

\hypersetup{
    colorlinks=true, % Enable colored links
    linkcolor=,  % Default color for internal links (sections, etc.)
    citecolor=,  % Default color for citations
    urlcolor=    % Default color for external URLs
}

%% file: appendix/dataset.tex
\begin{table*}[]
\centering
\begin{tabular}{@{}lcccc@{}}
\toprule
                                 & \textbf{Number} & \textbf{Number+Text} & \textbf{Easy Entity} & \textbf{Hard Entity} \\ \midrule
Count                            & 900             & 743                  & 900                  & 900                  \\
Average Answer Length (Tokens)   & 1.00            & 2.49                 & 2.77                 & 5.18                 \\
Average Question Length (Tokens) & 8.75            & 21.9                 & 18.9                 & 22.9                 \\ \bottomrule
\end{tabular}
\caption{Dataset details of each split (Number, Number+Text, Easy Entity, Hard Entity), including the number of data instances, average length of answers (in tokens), and average length of questions (in tokens). Tokens are computed using \texttt{tiktoken}.}
\label{table:dataset}
\end{table*}

%% file: appendix/metric_eval.tex
\begin{table*}[]
\centering
\begin{tabular}{@{}lcccc@{}}
\toprule
\multicolumn{1}{c}{\textbf{Metric}} & \textbf{Number} & \textbf{Number + Text} & \textbf{Easy Entity} & \textbf{Hard Entity} \\ \midrule
DSPy (GPT-4o) & 0.972 & 0.924 & \textbf{0.917} & \textbf{0.897} \\
Rule-Based & \textbf{0.979} & \textbf{0.965} & 0.817 & 0.790 \\
Exact Match & \textbf{0.979} & 0.819 & 0.752 & 0.537 \\
Token F1 & 0.969 & 0.771 & 0.845 & 0.829 \\
Token Recall & 0.969 & 0.760 & 0.848 & 0.826 \\
Token Precision & 0.969 & 0.760 & 0.848 & 0.826 \\
\textsc{PEDANTS} & 0.972 & 0.760 & 0.872 & 0.786 \\ \bottomrule
\end{tabular}
\caption{Raw agreement with human annotators (i.e. accuracy) of seven tested answer equivalence metrics. The best metric for each dataset split is in \textbf{bold}.}
\label{table:metrics}
\end{table*}

%% file: appendix/benchmark_human.tex
\begin{figure*}
    \centering
    \includegraphics[width=\linewidth]{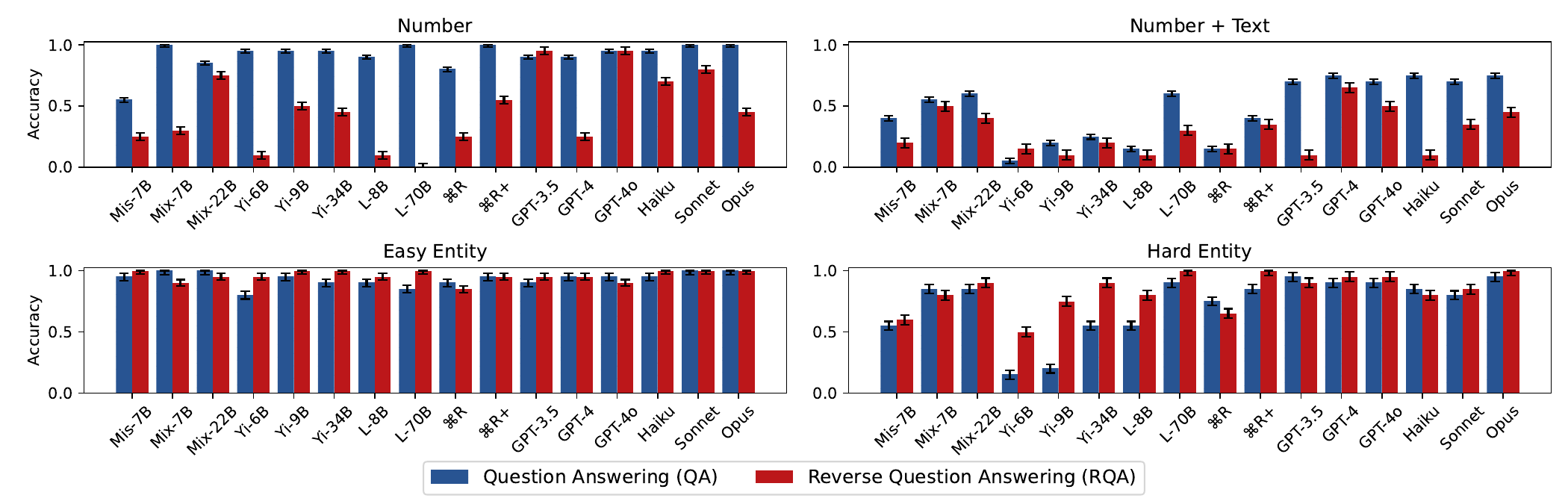}
    \vspace{-2.25ex}
        
        \caption{\mm{} deduction (\textbf{\textcolor{blue}{blue}}) and abduction (\textbf{\textcolor{red}{red}}) accuracy based on human annotations on a subset of data (20 labels per model/dataset). The plot shows a similar trend as the automated metrics (\mm{}s are weaker in abduction in numerical settings, but stronger in abduction in non-numerical settings), confirming the validity of our metrics.}
    \label{fig:benchmark_human}
    \vspace{-1ex}
\end{figure*}

%% file: appendix/prompt_eng.tex
\begin{figure*}
    \centering
    \includegraphics[width=\linewidth]{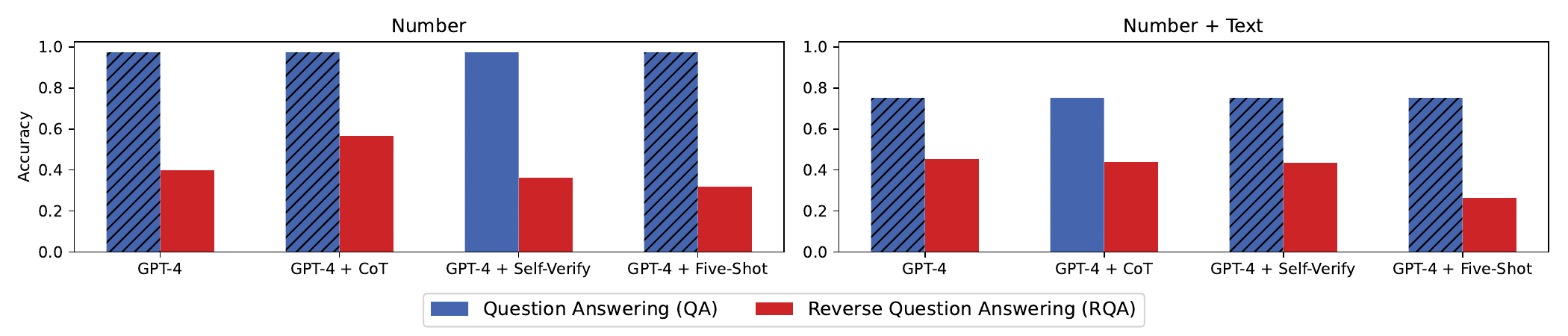}
        \caption{\mm{} deduction (\textbf{\textcolor{blue}{blue}}) and abduction (\textbf{\textcolor{red}{red}}) accuracy with GPT-4 on numerical entities. For \fwd, we present the zero-shot prompt used in \cref{subsection:accuracy}. For \bwd, we test adding chain-of-thought instructions (GPT-4 + CoT), asking the \mm{} to verify its question post-generation (GPT-4 + Self-Verification), and including five exemplars (GPT-4 + 5-Shot). None of these strategies allow the model to fully match the \fwd accuracy.}
    \label{fig:benchmark_prompt}
\end{figure*}

%% file: appendix/truth_table.tex
\begin{table*}[]
\small
\centering
\begin{tabular}{@{}cccc@{}}
\toprule
Is $a_{true}$ the answer to $q_{bwd}$? & Is $a_{bwd}$ the answer to $q_{bwd}$? & Is $a_{true}$ equal to $a_{bwd}$? & \textbf{Outcome} \\ \midrule
\texttt{Yes} & \texttt{Yes} & \texttt{Yes} & \bwd = \fwd \\
\texttt{Yes} & \texttt{Yes} & \texttt{No} & Ambiguous Question \\
\texttt{Yes} & \texttt{No} & \texttt{Yes} & \fwd Fails \\
\texttt{Yes} & \texttt{No} & \texttt{No} & Metric Error (Impossible) \\
\texttt{No} & \texttt{Yes} & \texttt{Yes} & \bwd Fails \\
\texttt{No} & \texttt{Yes} & \texttt{No} & Metric Error (Impossible) \\
\texttt{No} & \texttt{No} & \texttt{Yes} & \bwd + \fwd Fail \\
\texttt{No} & \texttt{No} & \texttt{No} & Mistakes Cancel (lucky!) \\ \bottomrule
\end{tabular}
\caption{All truth table outcomes for the consistency analysis in \cref{subsection:consistency}.}
\label{table:truth_table}
\end{table*}

%% file: appendix/consistency_all.tex
\begin{figure*}
    \centering
    \includegraphics[width=\linewidth]{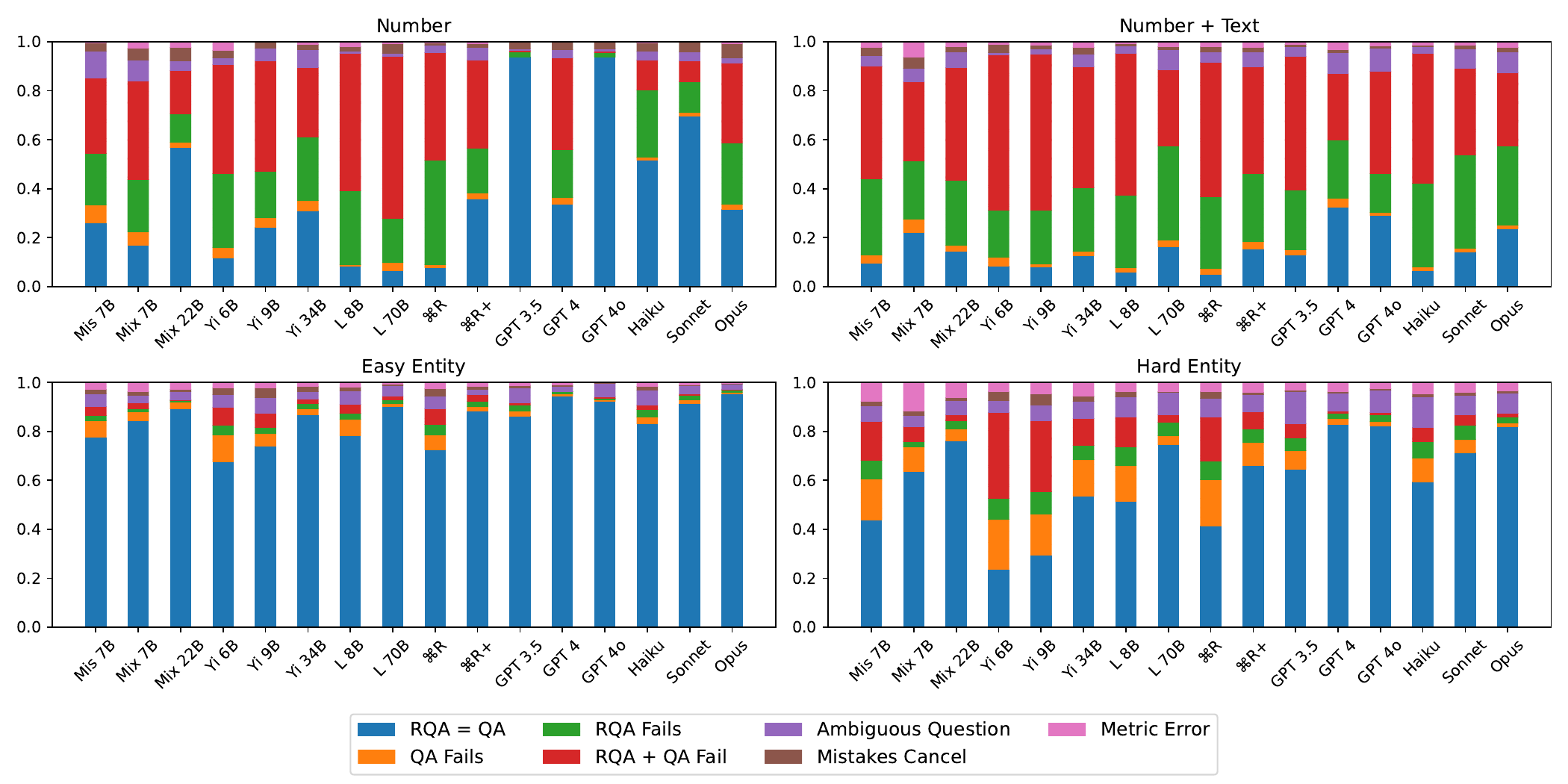}
    \vspace{-2.25ex}
        \caption{\fwd and \bwd logical consistency across all models. The consistency trends are also prevalent for smaller/less capable \mm{}s; \bwd and \fwd consistency is higher for easy/hard entities, but \mm{}s can often detect their own \bwd inaccuracies in numerical settings. }
    \label{fig:consistency_all}
    \vspace{-1ex}
\end{figure*}

%% file: appendix/errors_numtext.tex
\begin{figure*}
    \centering
    \includegraphics[width=\linewidth]{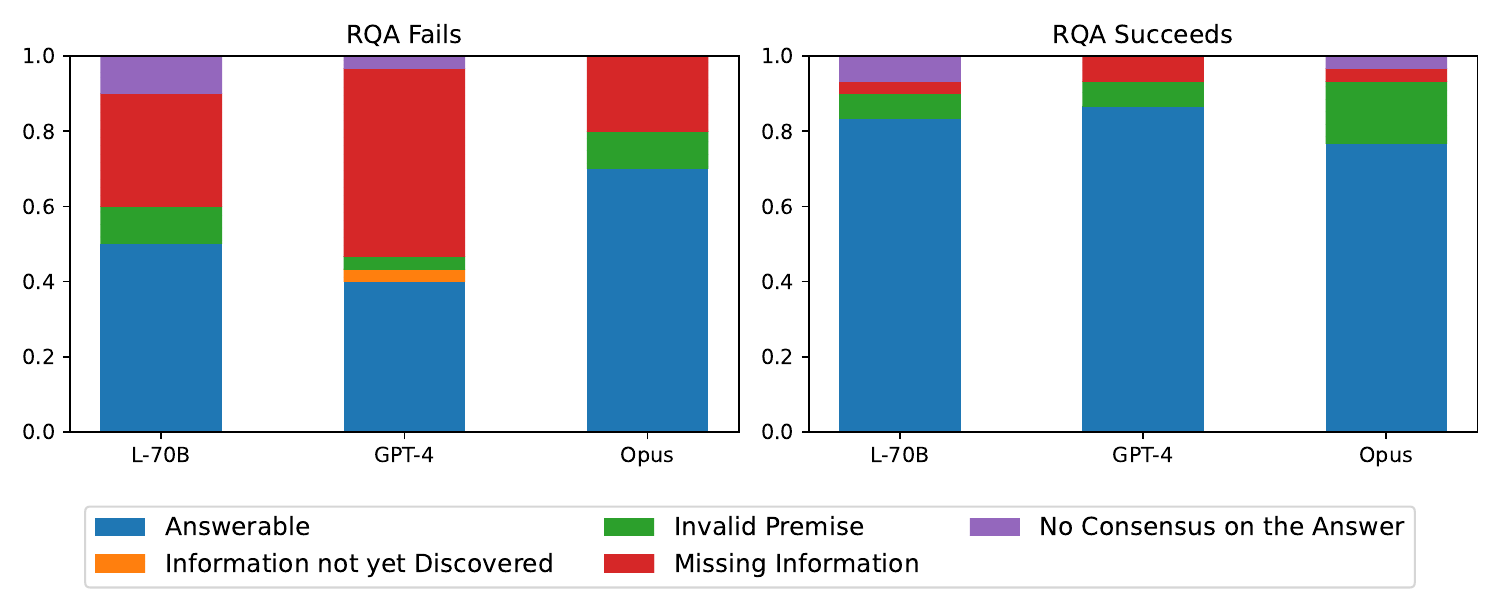}
    \caption{Error analysis of questions from \bwd on Number+Text. When \bwd fails, questions are often unanswerable (30-60\%), and frequently include false premises or omit key information that is needed to answer the question.}
    \label{fig:errors_numtext}
\end{figure*}

%% file: appendix/error_qual.tex
\begin{table*}[]
\scriptsize
\centering
\resizebox{\textwidth}{!}{\begin{tabular}{@{}lllll@{}}
\toprule
\textbf{Question} & \textbf{Answer} & \textbf{Model} & \multicolumn{1}{l}{\textbf{Valid?}} & \textbf{Question Type} \\ \midrule
What is the sum of the numbers on a standard roulette wheel? & 369 & L-70B & No & Multi-Step \\
What is the sum of the first 37 natural numbers? & 749 & GPT-4 & No & Multi-Step \\
What is the sum of the first 18 positive odd integers? & 855 & Opus & No & Multi-Step \\
What is the result of multiplying 25 by 25? & 625 & L-70B & Yes & Single-Step \\
What is the smallest prime number greater than 357? & 359 & GPT-4 & Yes & Single-Step \\
What is the product of 30 and 23? & 690 & Opus & Yes & Single-Step \\
What is the emergency telephone number in the United States and many other countries? & 911 & L-70B & Yes & Fact-based \\
What is the atomic number of the element with the highest atomic number ... as of 2023? & 223 & GPT-4 & No & Fact-based \\
What is the number of characters allowed in a single tweet on Twitter? & 280 & Opus & Yes & Fact-based \\ \bottomrule
\end{tabular}}
\caption{Examples of \bwd question types and errors on the Number split.}
\label{table:error_num_qual}
\end{table*}
\begin{table*}[]
\scriptsize
\centering
\resizebox{\textwidth}{!}{\begin{tabular}{@{}llll@{}}
\toprule
\textbf{Question} & \textbf{Answer} & \textbf{Model} & \textbf{Error Type} \\ \midrule
How many British soldiers were killed or wounded during the Battle of Thermopylae in 480 BCE? & 266 men & L-70B & Invalid Premise \\
What is the numerical designation..., if we humorously assume there were 111 before it? & 112 Ark & GPT-4 & Invalid Premise \\
According to a 2011 census, how many officially recognized ethnic groups are there in India? & \multicolumn{1}{l}{634 distinct peoples} & Opus & Invalid Premise \\
In what year did the Vietnamese king Le Hoan defeat the Song Dynasty army at the Battle of Bach Dang? & 988 AD & L-70B & No Consensus \\
How long did the construction of the Great Wall of China continue...? & 264 years & GPT-4 & No Consensus \\
What is the wavelength of yellow light in the visible spectrum? & 587 nanometers & L-70B & Missing Info \\
How many individuals attended the annual community festival last year according to the final headcount? & 178 people & GPT-4 & Missing Info \\
How old was the world's oldest tortoise, Jonathan, when he passed away in 2022? & 179 years of age & Opus & Missing Info \\ \bottomrule
\end{tabular}}
\caption{Examples of \bwd question types and errors on the Number+Text split.}
\label{table:error_num_text_qual}
\end{table*}

%% file: appendix/match.tex
\begin{figure*}
    \centering
    \includegraphics[width=\linewidth]{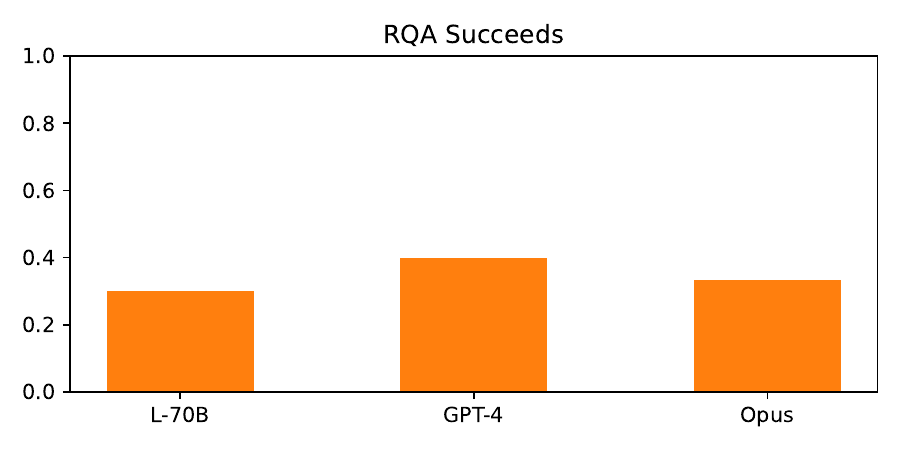}
    \caption{Proportion of \bwd questions on Numbers+Text that semantically match the ground-truth question when \bwd succeeds. LLaMA-3 70B, GPT-4, and Opus can all match the ground-truth question over 25\% of the time.}
    \label{fig:errors_match}
\end{figure*}

%% file: appendix/overlap.tex
\begin{figure*}
    \centering
    \includegraphics[width=\linewidth]{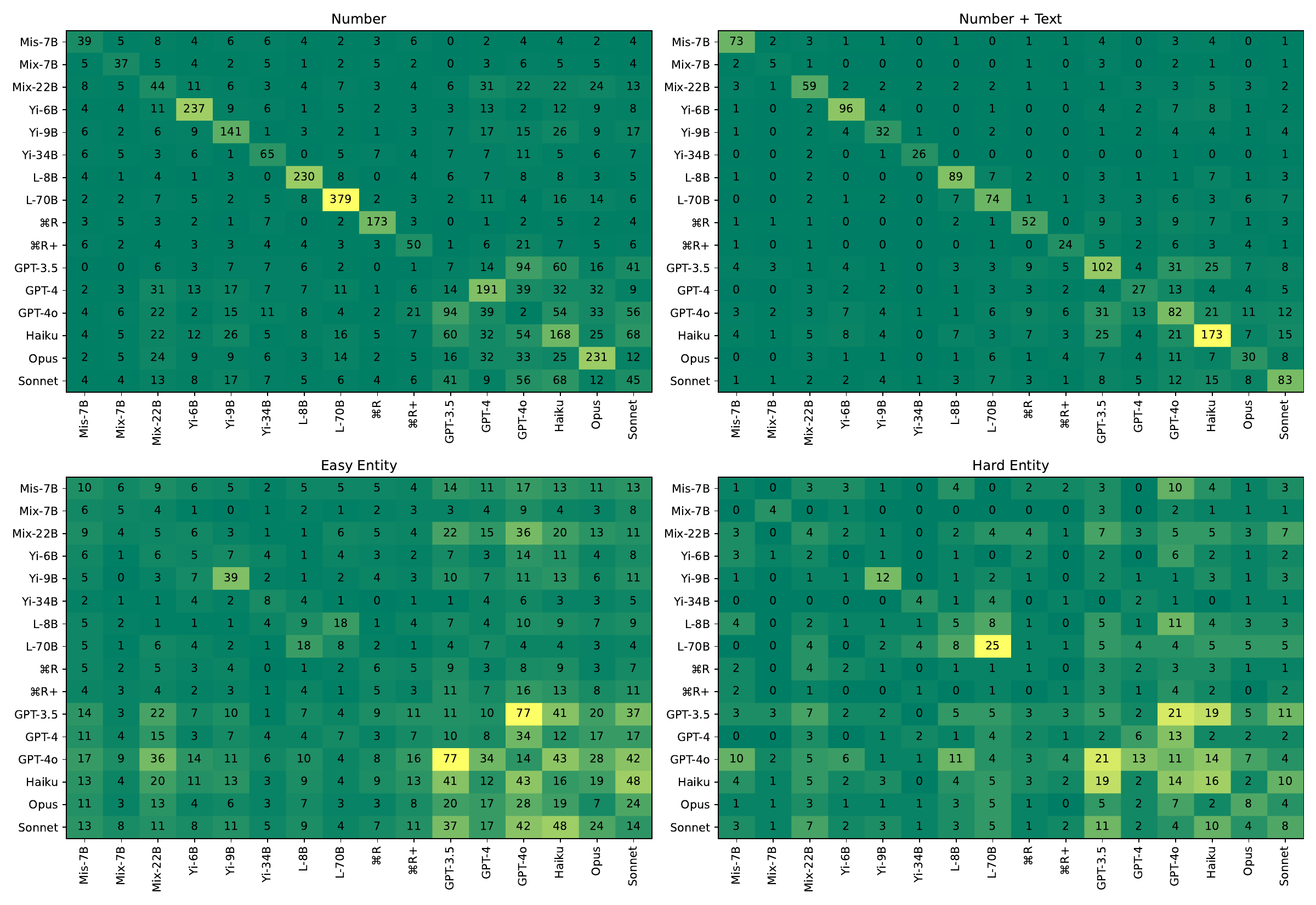}
    \caption{Cross-model frequency of questions from \bwd that are exact duplicates. \mm{}s often generate the same question in \bwd even though the input answer changes, reaching as high as 379 for LaMA-3 70B on Numbers.}
    \label{fig:overlap}
\end{figure*}

%% file: appendix/question_mem.tex
\begin{table*}[]
\centering
\begin{tabular}{lcccc|c}
\toprule
Model & Easy Fact & Hard Fact & Number & Number+Text & \textbf{Model Sum} \\
\midrule
Mis-7b & 25 & 6 & 6 & 7 & 44 \\
Mix-7B & 7 & 0 & 10 & 1 & 18 \\
Mix-22B & 41 & 3 & 17 & 12 & 73 \\
Yi-6B & 18 & 0 & 98 & 40 & 156 \\
Yi-9B & 17 & 2 & 8 & 1 & 28 \\
Yi-34B & 7 & 0 & 18 & 0 & 25 \\
L-8B & 12 & 2 & 21 & 8 & 43 \\
L-70B & 4 & 1 & 19 & 4 & 28 \\
Command-R & 48 & 3 & 61 & 47 & 159 \\
Command-R+ & 28 & 3 & 17 & 20 & 68 \\
GPT-3.5 & 111 & 19 & 16 & 105 & 251 \\
GPT-4 & 29 & 0 & 32 & 3 & 64 \\
GPT-4o & 96 & 10 & 27 & 39 & 172 \\
Haiku & 88 & 12 & 67 & 95 & 262 \\
Sonnet & 51 & 5 & 18 & 17 & 91 \\ \midrule
Opus & 41 & 0 & 48 & 12 & 101 \\
\textbf{Dataset Sum} & 623 & 66 & 483 & 411 & \textbf{1583} \\
\bottomrule
\end{tabular}
\caption{Number of generated \bwd questions that are exact matches to a question in the Dolma pretraining corpus. On average, models are prone to copying questions from pretraining $\sim3$\% of the time. Smaller/weaker \mm{}s are more susceptible to copying questions from pretraining in \bwd. Further, easy facts and numerical answers are more likely to lead to copied questions in \bwd versus our hard facts.
}
\label{table:question_memorization}
\end{table*}

%% file: appendix/memorization_ex.tex
\begin{table*}[]
\centering
\setlength{\tabcolsep}{3pt}
\resizebox{\textwidth}{!}{\begin{tabular}{@{}llllll@{}}
\toprule
\textbf{Question} & \textbf{Answer} & \textbf{Model(s)} & \textbf{Split} & \textbf{Valid} & \textbf{Count} \\ \midrule
What is the answer to this question? & Lucy poems & Haiku & Hard Fact & No & 21313 \\
Who lives in a pineapple under the sea? & Spongebob Squarepants & GPT-3.5, GPT-4o & Easy Fact & Yes & 1452 \\
Where does the story take place? & In the Penal Colony & GPT-3.5 & Hard Fact & No & 1395 \\
How many countries are there in the world? & 195 nations & GPT-3.5 & Num+Text & Yes & 380 \\
What is the capital of France? & Paris, France & Command-R+ & Easy Fact & Yes & 338 \\
Who was the first president of the United States? & George Washington & Haiku, Sonnet & Easy Fact & Yes & 281 \\
How many days are there in a week? & 357 & Yi-6B & Number & No & 194 \\
What is the capital of the United States? & Washington, D.C. & Command-R, GPT-3.5 & Easy Fact & Yes & 192 \\
How many days are there in a year? & 365 & Command-R & Easy Fact & Yes & 166 \\
How many days are there in a year? & 800 & Haiku & Number & No & 166 \\ \bottomrule
\end{tabular}}
\caption{Questions generated from \bwd that are most frequently found in the Dolma corpus. The \mm{}'s tendency to generate inaccurate questions (e.g. \textit{How many days are there in a year?} for \textit{800}) or ambiguous questions (\textit{What is the answer to this question?}) could be influenced by how often these questions appear in pretraining.}
\label{table:memorization}
\end{table*}